\RequirePackage{fix-cm}
\documentclass[natbib, extended, 16pt]{svjour3}       % onecolumn (second format)
\smartqed  % flush right qed marks, e.g. at end of proof
\usepackage{graphicx}

\usepackage{mathptmx}      % use Times fonts if available on your TeX system

\usepackage{amsmath, amssymb}
\usepackage{bm}
\usepackage{mathrsfs}
\usepackage{graphicx, fancyhdr}
\usepackage{placeins}  % prevent unnecessary floating
\usepackage{floatrow}

\DeclareMathOperator*{\argmax}{arg\,max}

\usepackage[vlined,linesnumbered,boxruled]{algorithm2e}
\usepackage{epsfig}
\usepackage[intoc]{nomencl}
\usepackage{hyperref}

\let\origfontsize\fontsize
\def\fontsize#1#2{\origfontsize{12}{14.5}}

\hypersetup{
 colorlinks=true,
 linkcolor=blue,
 citecolor=blue,
 urlcolor=blue
 }

\usepackage{epstopdf}
\usepackage{color}
\usepackage[boxruled,vlined,linesnumbered]{algorithm2e}
\let\oldnl\nl% Store \nl in \oldnl
\newcommand{\nonl}{\renewcommand{\nl}{\let\nl\oldnl}}% Remove line number for one line

\usepackage{ifpdf}
\usepackage{multirow}
\usepackage{caption}
\usepackage{subcaption}

\newdimen\figrasterwd
\figrasterwd\textwidth

\usepackage{setspace}
\usepackage{xcolor}

\usepackage[margin=1in]{geometry}

\newlength\myindent
\journalname{my journal}
\begin{document}

\title{Deep Reinforcement Learning for Constrained Field Development Optimization in Subsurface Two-phase Flow}

\author{Yusuf~Nasir$^{2}$ \and Jincong~He$^{1}$ \and Chaoshun~Hu$^{1}$ \and  Shusei~Tanaka$^{1}$ \and Kainan~Wang$^{1}$ \and XianHuan~Wen$^{1}$}

\authorrunning{Y.~Nasir, J.~He, C.~Hu, S.~Tanaka, K.~Wang, X.~Wen} % if too long for running head

\institute{Corresponding author: Yusuf~Nasir \at
          \email{nyusuf@stanford.edu}
 \\ \\
            $^{1}$ \ Chevron ETC \\ \\
            $^{2}$ \ Stanford University
}

\date{Received: date / Accepted: date}
\titlerunning{DRL for Constrained FDO}

\maketitle

\begin{abstract}

Oil and gas field development optimization, which involves the determination of the optimal number of wells, their drilling sequence and locations while satisfying operational and economic constraints, represents a challenging computational problem. In this work, we present a deep reinforcement learning-based artificial intelligence agent that could provide optimized development plans given a basic description of the reservoir and rock/fluid properties with minimal computational cost. This artificial intelligence agent, comprising of a convolutional neural network, provides a mapping from a given state of the reservoir model, constraints, and economic condition to the optimal decision (drill/do not drill and well location) to be taken in the next stage of the defined sequential field development planning process. The state of the reservoir model is defined using parameters that appear in the governing equations of the two-phase flow (such as well index, transmissibility, fluid mobility, and accumulation, etc.).

A feedback loop training process referred to as deep reinforcement learning is used to train an artificial intelligence agent with such a capability. The training entails millions of flow simulations with varying reservoir model descriptions (structural, rock and fluid properties), operational constraints (maximum liquid production, drilling duration, and water-cut limit), and economic conditions. The parameters that define the reservoir model, operational constraints, and economic conditions are randomly sampled from a defined range of applicability. Several algorithmic treatments are introduced to enhance the training of the artificial intelligence agent. After appropriate training, the artificial intelligence agent provides an optimized field development plan instantly for new scenarios within the defined range of applicability. This approach has advantages over traditional optimization algorithms (e.g., particle swarm optimization, genetic algorithm) that are generally used to find a solution for a specific field development scenario and typically not generalizable to different scenarios. The performance of the artificial intelligence agents for two- and three-dimensional subsurface flow are compared to well-pattern agents. Optimization results using the new procedure are shown to significantly outperform those from the well pattern agents.

Compared to prior work on this topic such as~\citet{He2021Deep}, the novelties in this work include:
\begin{itemize}
    \item Extended the problem description from single-phase flow to two-phase flow, and thus allowing for handling of reservoirs with strong aquifers and waterflooding problems.
    \item Redesigned deep network neural network architecture for better performance.
    \item Extended to handle operational constraints such as maximum liquid production, drilling duration and water-cut limit.
    \item Extended to three-dimensional reservoir model.
\end{itemize}

% \keywords{Field Development Optimization \and Well-Pattern Optimization \and Well Placement \and Drilling Schedule \and Realization-by-Realization Well Completion}
\end{abstract}

%===============================================================================
\section{Introduction}
\label{sec-intro}
%===============================================================================

Field development decisions such as the number of wells to drill, their location and drilling sequence need to be made optimally to maximize the value realized from a petroleum asset. Optimization algorithms such as evolutionary strategies are, in recent times, widely applied to the field development optimization problem~\citep{isebor2014a, isebor2014b}. However, the application of these optimization algorithms is challenging due to the very large number of computationally expensive flow simulations required to obtain optimal (or near optimal) solutions. In addition, the field development optimization problem is typically solved separately for each petroleum field due to the variation in geological model, constraints to be considered or even the economic condition. Thus, the large number of computationally expensive flow simulations needs to be run for each field under consideration. This suggests the field development optimization problem will benefit from strategies that would allow for the generalization of the optimization process to several petroleum fields.

In our recent work \cite{He2021Deep}, we developed a deep reinforcement learning technique for the field development optimization in two-dimensional subsurface single-phase flow settings. The deep reinforcement learning technique allows for the training of an artificial intelligence agent that provides a mapping from the current state of a two-dimensional reservoir model to the optimal decision (drill/do not drill and well location) in the next step of the development plan. Our goal in this work is to extend the procedures in \cite{He2021Deep} for the field development optimization, in the presence of operational constraints, in both two- and three-dimensional subsurface two-phase flow. Once properly trained, the artificial intelligence agent should learn the field development logic and provide optimized field development plans instantly for different field development scenarios. Besides the described extensions to more complex simulation models, several algorithmic treatments are also introduced in this work to enhance the training of the artificial intelligence agent under the deep reinforcement learning framework. 

In the literature of oil and gas field development and production optimization, different optimization algorithms have been applied to solve different aspects of the optimization problem. The well control or production optimization problem in which the time-varying operational settings of existing wells are optimized has been efficiently solved with gradient-based methods~\citep{brouwer2002dynamic, sarma2006efficient,echeverria2011application, awotunde2019comprehensive}, ensemble-based methods~\citep{fonseca2014ensemble, chen2009efficient} or with efficient proxies like reduced-order models~\citep{cardoso2010linearized, he11a}. For the well placement optimization, popular algorithms such as the genetic algorithms and the particle swarm optimization algorithms~\citep{bangerth2006optimization, onwunalu2010application, bouzarkouna2012well, isebor2014a, isebor2014b} typically entail thousands of simulation runs to get improved result with no guarantee of global optimality. The joint well placement and production optimization problems have also been considered~\citep{zandvliet2008adjoint, bellout2012joint, isebor2014b, isebor2014a, nasir2019hybrid}, which typically require even larger numbers of simulations. In addition, while the approaches considered in these studies may provide satisfactory results for the field development optimization problem, the solution obtained in each case is tied to a specific field development scenario. If the economic conditions or geological models used in the optimization change, the optimization process needs to be repeated. In other words, the solution from traditional optimization methods lacks the ability to generalize when the underlying scenario changes. It should be noted that the ability to generalize that is discussed here is different from the robust optimization as studied in~\citet{chen2012robust, chen2017minimizing}, in which the optimization is performed under uncertainty. While the solutions from robust optimization accounts for the uncertainty in the the model parameters, they still don't generalize when the ranges of those uncertainties change. 

In this work, we consider the reinforcement learning technique in which a general AI (in the form of a deep neural network) can be applied to optimize field development planning for a range of different scenarios (e.g., different reservoir, different economics, different operational constraints) once it is trained. The reinforcement learning technique considered in this work for the field development optimization problem has shown great promise in other fields. Google DeepMind trained an artificial intelligence agent AlphaGo~\citep{silver2016mastering} using reinforcement learning with a database of human expert games. AlphaGo beat the human world champion player in the game of Go. AlphaGo Zero~\citep{AlphaGoZero2017} which is a variant of AlphaGo was trained through self-play without any human knowledge. AlphaGo Zero defeated AlphaGo in the game of Go. In AlphaGo and AlphaGo Zero, the AI takes in a description of the current state of the Go game and chooses an action to take for the next step. The choice of action is optimized in the sense that it maximizes the overall probability of winning the game. The success of AlphaGo and AlphaGo Zero suggests the possibility of training an AI for field development that takes in a description of the current state of the reservoir and chooses the development options (e.g., drilling actions) for the next step, without prior reservoir engineering knowledge. 

In the petroleum engineering literature, deep learning algorithms have seen much success recently for constructing proxies for reservoir simulation models~\citep{wang2021efficient, jin2020deep, tang2021deep}. Reinforcement learning algorithms have been applied to solve the production optimization problem for both steam injection in steam-assisted gravity drainage (SAGD) recovery process~\citep{guevara2018optimization} and waterflooding~\citep{hourfar2019reinforcement}. Deep reinforcement learning (in which the artificial intelligence agent is represented by a deep neural network) has also been applied to the production optimization problem~\citep{ma2019waterflooding, miftakhov2020deep}. \cite{ma2019waterflooding} evaluated the performance of different deep reinforcement learning algorithms on the well control optimization problem with fully connected neural networks (FCNN). The state of the reservoir model defined by two-dimensional maps are flattened and used as input to the FCNN and thus do not retain the spatial information inherent in the data. An FCNN-based artificial intelligence agent was also trained in \cite{miftakhov2020deep}. It should be noted that in the studies discussed here, the reinforcement learning process is used as a replacement for traditional optimization algorithms (such as  particle swarm optimization, genetic algorithm) to optimize a predefined field development scenario.

\cite{He2021Deep} applied the deep reinforcement learning technique to the more challenging field development optimization problem in which the decisions to be made include the number of wells to drill, their locations, and drilling sequence. In contrast to the previous studies discussed, \cite{He2021Deep} used DRL to develop artificial intelligence which could provide optimized field development plans given any reservoir description within a predefined range of applicability. In addition, unlike in previous studies~\citep{ma2019waterflooding, miftakhov2020deep} in which the artificial intelligence agent is represented by fully connected neural networks, \cite{He2021Deep} utilized a convolutional neural network to represent the agent which allows for better processing of spatial information. The field development optimization problem in \citet{He2021Deep}, however, considers a single-phase flow in two-dimensional reservoir models.

In this paper, we build upon~\citet{He2021Deep} and formulate deep reinforcement learning-based artificial intelligence agents that could provide optimized field development plans instantaneously based on the description of a two- or three-dimensional subsurface two-phase flow in the presence of operational constraints. The sequence of actions to take during the field development process is made by the artificial intelligence agent after processing the state of the reservoir model, the prescribed constraints and economic condition. The state of the reservoir model in this work is defined using parameters that appear in the governing equations of the two-phase flow (such as pressure, saturation, well index, transmissibility, fluid mobility, and accumulation). We also propose a dual-action probability distribution parameterization and an improved convolutional neural network architecture to enhance the training efficiency of the artificial intelligence agents for the field development optimization problem.

This paper proceeds as follows. In Section~\ref{sec:tr_rl_fdo}, we present the governing equations for the two-phase flow and discuss the different field development optimization approaches -- traditional and reinforcement learning-based field development optimization. In Section~\ref{sec:drl_fdo}, we present the deep reinforcement learning field development approach which includes the state and action representation, the proximal policy algorithm and the deep neural network used to represent the policy and value functions of the agent. Computational results demonstrating the performance of the deep reinforcement learning agents, for both 2D and 3D problems, are presented in Section~\ref{sec:results}. We conclude in Section~\ref{sec:concl} with a summary and suggestions for future work.

%===============================================================================
\section{Governing Equations and Field Development Optimization Approaches}
\label{sec:tr_rl_fdo}
%===============================================================================
In this section, we first briefly discuss the governing equations for the two-phase flow. We then describe the traditional and reinforcement learning-based field development optimization approaches.

\subsection{Governing equations}

In this work, we consider the isothermal immiscible oil-water flow problem with gravitational effects. Combining Darcy's law for multiphase flow and the mass conservation equation while neglecting capillary pressure, the flow of each phase in the reservoir can be described using:
\begin{equation}
    \nabla \cdot \Big[ \textbf{k}\rho_{l} \lambda_{l} \left(\nabla p - \gamma_{l} \nabla D \right) \Big] = \frac{\partial }{\partial t}  \left(\phi \rho_{l} S_{l}\right) + q_{l},
    \label{Eqn:2p_eqn}
\end{equation}

\noindent 
where the subscript $l$ represents the phase ($l = o$ for oil and $l = w$ for water), \textbf{k} is the permeability tensor, $\rho_{l}$ is the phase density, $\lambda_{l} = k_{r,l}/\mu_{l}$ is the phase mobility, with $k_{r,l}$ the phase relative permeability and $\mu_{l}$ the phase viscosity, $p$ is the pressure (with $p = p_{o} = p_{w}$ since capillary pressure is neglected), $\gamma_{l} = \rho_{l}g$ is the phase specific weight, with $g$ the gravitational acceleration, $D$ is the depth, $\phi$ is the porosity, $S_{l}$ is the phase saturation and $q_{l}$ is the mass sink term. The phase flow rate for a production well $w$ in well-block $i$ is defined by the Peaceman well model~\citep{peaceman1983interpretation}:

\begin{equation}\label{Eqn:rate}
    \left(q_{l}^{w}\right)_{i} = WI_{i} \left(\lambda_{l} \rho_{l} \right)_{i} (p_{i} - p^{w}),
\end{equation}
where $WI_{i}$ is the well-block well index which is a function of the well radius, well block geometry and permeability, $p_{i}$ denotes the well block pressure and $p^w$ denotes the well bottomhole pressure.

The discretized form of Eq.~\ref{Eqn:2p_eqn} given in Eq.~\ref{Eqn:discretized_equation} is solved for each grid block $i$, in time, using the fully implicit method.

\begin{equation}\label{Eqn:discretized_equation}
    \left[\frac{(c_t)_l\phi V S_l}{B_l}\right]_i\frac{p_i^{n+1}-p_i^{n}}{\Delta t} = \sum\limits_{k}^{} \Gamma_{l}^{k} \Delta \Psi_{l}^{k} + (q_{l})_i,
\end{equation}

\noindent 
where $p_i$ the pressure of grid block $i$, $n$ and $n + 1$ indicate the time levels, $V$ is the volume of the grid block, $(c_t)_l$ denotes the total phase compressibility of the grid block, $B_l$ is the phase formation volume factor of the grid block, $k$ represents an interface connected to grid block $i$, $\Delta  \Psi_{l}^{k}$ is the difference in phase potential over the interface $k$, $\Gamma_{l}^{k} = \Gamma^{k} \lambda_{l}^{k}$ is the phase transmissibility over the interface $k$, with $\Gamma^{k}$ the rock transmissibility which is a function of the permeability and geometry of the grid blocks connected by interface $k$. 

\subsection{Optimization problem formulation and traditional approaches}

The aim in the field development optimization problem in this work is to determine the number of wells to drill in a greenfield, alongside their locations and drilling sequence. Only the primary depletion mechanism is considered, thus all wells to be drilled are production wells. The constraints imposed on the field development include the maximum liquid production rate, drilling duration for each well, the allowable water cut limit, and minimum inter-well spacing.

Mathematically, the field development optimization problem can be written as follows: 
\begin{gather}
\begin{array}{rrclcl}
\displaystyle \max_{\textbf{x} \in \mathbb{X}} & {J (\textbf{x})}, \ \  \textrm{subject to } \textbf{c}(\textbf{x})\leq\textbf{0},
\end{array}
\label{eq:trad_field_dev_opt_eqn}
\end{gather}
\noindent 
where $J$ is the objective function to be optimized, the decision vector $\textbf{x} \in \mathbb{X}$ defines the number of wells to drill, their locations, and drilling sequence. The space $\mathbb{X}$ defines the feasible region (upper and lower bounds) for the decision variables. The vector \textbf{c} defines optimization constraints that should be satisfied. Interested readers can refer to ~\cite{isebor2014b, isebor2014a} for possible ways of parameterizing the decision vector $\textbf{x}$ for the field development optimization problem.

The net present value (NPV) is typically considered as the objective function to be maximized in the field development optimization. Thus, $J$ in Eq.~\ref{eq:trad_field_dev_opt_eqn} can be specified as the NPV. The NPV for primary depletion (only production wells) can be computed as follows:
\begin{equation}
	\textnormal{NPV(\textbf{x})}=\sum\limits_{t = 0}^{T}  \gamma^{t_k}\left[ \sum\limits_{i = 1}^{N_{w,p,t}} \left(\left(p_{o} - c_{opex}\right) ~q^{i}_{o,k}-c_{pw}~q^{i}_{w,k}\right) - \sum\limits_{i = 1}^{N_{w,d,t}}  c_w \right],
    \label{Eqn:gen_field_dev_npv_eqn}
\end{equation} 
\noindent 
Here $N_t$ is the number of time steps in the flow simulation, $N_w$ is the number of production wells, $t_k$ and $\Delta t_k$ are the time and time step size at time step $k$, $t_i$ is the time at which producer $i$ is drilled, and $p_{o}$, $c_{opex}$ and $c_{pw}$ represent the oil price, operating cost and the cost of produced water, respectively. The variables $c_w$ and $\gamma$ represent the well drilling cost and annual discount rate, respectively. The rates of oil and water production for well $i$ at time step $k$ are, respectively, $q^{i}_{o,k}$ and $q^{i}_{w,k}$.

In the traditional approach of solving the field development optimization problem, the field development scenario of interest is first defined. These include the reservoir model definition (structural, rock, and fluid properties), the specific constraints and the economic condition to be considered. Afterward, the optimization variables that define the number of well, location, and drilling sequence are parameterized to obtain a formulation of $\textbf{x}$. Finally, an optimizer (e.g. particle swarm optimization or genetic algorithm) iteratively proposes sets of decision variables (that represent field development plans and are evaluated with a reservoir simulator) in order to maximize an economic metric of interest. The optimized set of decision variables is tied to a particular field development scenario predefined before the flow simulation.

In the traditional approach, the evaluation of Eq.~\ref{Eqn:gen_field_dev_npv_eqn} for each field development plan requires performing the two-phase flow simulation which involves solving Eq.~\ref{Eqn:discretized_equation}. From those simulations, the only information used by the traditional optimization approach is the corresponding NPV. The reservoir states at each time step, which contain a lot of useful information, are discarded.

Different from the traditional optimization approach, Reinforcement learning makes use of the intermediate states generated from the simulator and provides optimized policy applicable to a range of different scenarios. 

\subsection{Reinforcement learning-based field development optimization}

\begin{figure}[htbp]
	\centering
	\includegraphics[width=0.7\textwidth]{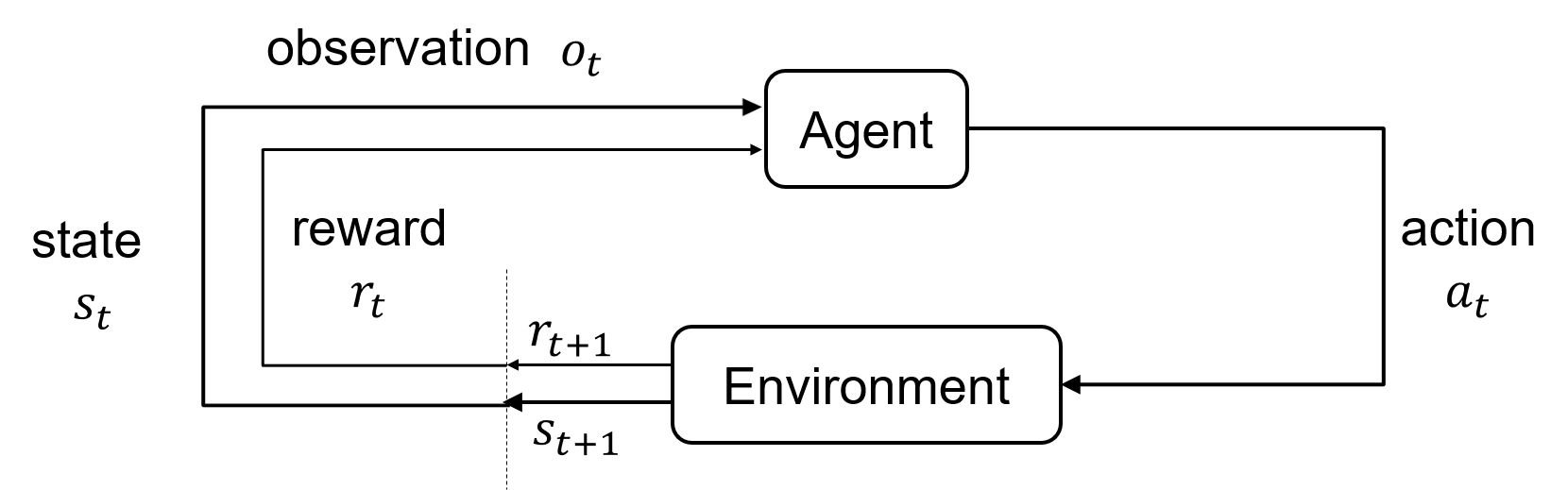}
	\caption{The reinforcement learning feedback loop \citep[adapted from][]{sutton2018reinforcement}.}
	\label{fig:rl_demo}
\end{figure}

Reinforcement learning is a sub-field of machine learning concerned with teaching an artificial intelligence agent how to make decisions so as to maximize the expected cumulative reward. The training and the decision making process of the AI agent follows a feedback paradigm as shown in Fig.~\ref{fig:rl_demo}.

The reinforcement learning problem can be expressed as a system consisting of an agent and an environment. The agent and the environment interact and exchange signals. These signals are utilized by the agent to maximize a given objective. The exchanged signals, also referred to as experience, are ($s_t, a_t, r_t$), which denotes the state, action, and reward respectively. Here, $t$ defines the time step in which the experience occurred. At a given stage $t$ of the decision-making process, with the environment (which represents where the action is taken) at state $s_t$, the agent takes an action or decision $a_t$. The quality of the action $a_t$ is quantified and signaled to the agent through the reward $r_t$. The decision-making process then transits to a new state $s_{t+1}$ which indicates the effect of action $a_t$ on the environment. The transition from $s_t$ to $s_{t+1}$ in reinforcement learning is formulated as a Markov decision process (MDP) -- which assumes the transition to the next state $s_{t+1}$  only depends on the previous state $s_t$ and the current action $a_t$. This assumption is referred to as the Markov property~\citep{howard1960dynamic}. The feedback loop terminates at a terminal state or a maximum time step $t = T$. The time horizon from $t = 0$ to when the environment terminates is referred to as an episode.

The choice of the action to take at stage $t$ by the agent depends on the observation $o_t$ which is a function of $s_t$. If $o_t$ contains all the information in $s_t$, then the decision making process is referred to as fully observable MDP. Otherwise, it is referred to a partially observable MDP (POMDP). The agent's action-producing function is referred to as a policy. Given a state $s_t$ (relayed to the agent through $o_t$), the policy produces the action $a_t$. This mapping operation is mathematically represented as $\pi(a_t \ | \ o_t, \ \theta)$, where $\theta$ are the parameters that define the policy function. The reinforcement learning problem is now essentially an optimization problem with the goal of finding the optimal policy parameters ($\theta^{opt}$) that maximizes an expected cumulative reward. This optimization problem is posed as:

\begin{equation}\label{Eqn:policy_opt}
  \theta^{opt} = \argmax_{\theta}{\mathbb{E} \left[\sum^{T}_{t=0} \gamma^t r_t\right]},
\end{equation}
where the reward at time $t$, $r_t = \mathbb{R}(s_t, a_t, s_{t+1})$ depends on the reward function $\mathbb{R}$ and $\gamma \in [0, 1]$ is the discount factor that accounts for the temporal value of rewards. The expectation accounts for the stochasticity that may exist in the action and environment. If the parameters of the policy function are the weights of a deep neural network, then the optimization problem is referred to as a deep reinforcement learning (described in details later). 

In the context of field development optimization, our goal is to determine the policy function that maximizes the expected NPV. At each drilling stage, the agent defines the action to be taken, which includes whether to drill a producer or not and, if drill, the optimal well location. The environment, which is the reservoir simulator, advances the flow simulation to the next drilling stage by solving the governing equation for the immiscible oil-water system given in Eq.~\ref{Eqn:2p_eqn}. The reward function $\mathbb{R}$ for the reinforcement learning-based field development optimization problem is given by the NPV~(Eq.~\ref{Eqn:gen_field_dev_npv_eqn}), where the reward for a drilling stage $t$ is given by $\textnormal{NPV}_t = \textnormal{NPV}(s_t, a_t, s_{t+1})$. Here, $\textnormal{NPV}(s_t, a_t, s_{t+1})$ defines the NPV obtained if the flow simulation starts from state $s_t$ (including all wells drilled from $s_0$ to $s_t$) and ends at state $s_{t+1}$. It should be noted that since the NPV defined in Eq.~\ref{Eqn:gen_field_dev_npv_eqn} accounts for discounting, we specify the the discount factor $\gamma = 1$ in Eq.~\ref{Eqn:policy_opt}.

The formulation of the RL-based field development optimization in this work allows the exploitation of the problem states ($s_t$) generated during the simulation and is thus a more efficient use of the information from the potentially expensive simulation runs. In addition, the output of the RL-based approach is not the optimal actions itself, but the optimal policy, a mapping from the states to the optimal actions $\pi(a_t|o_t,\theta)$, which can be used obtain the optimal actions under various different scenarios. This is distinctively different from the use of RL in~\citet{ma2019waterflooding, miftakhov2020deep}, where the goal was the optimal actions rather than the policy. 

%===============================================================================
\section{Deep Reinforcement Learning for Field Development Optimization}
\label{sec:drl_fdo}
%===============================================================================

In this section, we present the deep reinforcement learning approach for field development optimization where the parameters ($\theta$) of the policy function are defined by the weight of a convolutional neural network. We first describe the action and state representation for the field development optimization problem. Finally, the training procedure of the agent using the proximal policy optimization algorithm~\citep{schulman2017proximal} and the convolutional neural network architecture are described.

\subsection{Action representation}

The action representation is important because it affects the computational complexity of the learning process. At each drilling stage, the agent needs to decide to drill a well at a certain location or not drill at all. We introduce two variables to make these decisions. The drill or do not drill decision at drilling stage $t$ is represented by a binary categorical variable $w_{t} \in \{0,1 \}$ where 0 represents the decision not to drill a well and 1 the decision to drill. The second variable (active only when the agent decides to drill a well, i.e. $w_t = 1$) defines where the well should be drilled and is represented by the well location variable $u_t$, which is chosen from the possible drilling locations denoted by $\textbf{u} \in \mathbb{Z}^{N_xN_y}$ (the 2D grid flattened into 1D), where $N_x$ and $N_y$ represents the number of grid blocks in the areal $x$ and $y$ directions of the reservoir model. Each index in $\textbf{u}$ maps to a grid block in the reservoir model. Thus, at each drilling stage $t$, the action $a_t$ is defined by $a_t = [w_t, u_t]$, where $u_t$ is only considered if $w_t = 1$.

The action representation used in this work differs from that used in \cite{He2021Deep}. Specifically, in \cite{He2021Deep} a single variable of dimension $N_x \times N_y + 1$ is used to define the possible actions at each drilling stage, where the additional index represents the do-not-drill decision. Such an action parameterization means the action $a_t$ at each drilling stage is sampled from a single probability distribution of cardinality $N_x \times N_y + 1$. Thus, the do-not-drill decision depends on the number of possible well locations ($N_x \times N_y$). As the size of the reservoir model increases, it becomes increasingly difficult for the agent to learn when not to drill a well due to the cardinality of the probability distribution dominated by possible drilling locations. This is in contrast to the approach used in this work where the variables $w_t$ and $u_t$ are sampled from two different probability distributions of cardinality 2 and $N_x \times N_y$, respectively. Hence, the drill or do-not-drill action is independent of the size of the reservoir model. The parameterization used in this work can be naturally extended to other field development cases. For example, for waterflooding, the decision on the well type can be incorporated in $w_t$ as $w_{t} \in \{-1, 0,1 \}$, where -1 represents the decision to drill an injector, 0 to not drill a well, and 1 to drill a producer.

Given the vector $\textbf{u}$, drilling in some grid blocks may lead to an unacceptable field development plan from an engineering standpoint. These include field development plans where more than one well is to be drilled in the same location, violation of minimum inter-well distance constraint (acceptable spacing between wells) and/or drilling in inactive regions of the reservoir. At any given drilling stage, an action mask of the same dimension with $\textbf{u}$ defines acceptable drilling locations that the agent could choose from. Specifically, the probability of sampling an invalid drilling location in $\textbf{u}$ is set to zero. This action masking technique~\citep{huang2020closer, tang2020implementing} have been proven to be an effective strategy to improve the convergence of the policy optimization and also ensures the agent only takes valid action. The action masking technique was used in \cite{He2021Deep} for the field development optimization problem to ensure only feasible drilling locations are proposed by the agent.

\subsection{State representation}

The definition of the state of the environment for the field development optimization problem depends on the model physics. For example, the state would include pressure for single-phase flow and, additionally, saturation for two-phase flow. The definition of the state also depends on the kind of generalization capability that we want the AI to have. For example, if we want the AI to provide different optimal solutions for different geological models, geological structures and properties such as permeability, porosity should also be included in the state definition. If we want the AI to be able to provide different optimal solutions under different oil prices, the oil price should also be part of the state. 

Our goal in this work is to develop an AI for two-phase flow that can provide optimal solutions for different field development optimization scenarios with variable reservoir models, operational constraints, and economic conditions within a predefined range. Thus, this variation in the scenarios should be captured in the state representation.

In this work, the field development optimization scenarios considered are characterized by parameters following the distributions listed in Table~\ref{tab:range_of_param}. Each scenario defines a given geological structure, rock and fluid properties, operational constraints, and economic conditions. This includes the grid size, the spatial distribution of the grid thickness obtained through Sequential Gaussian simulation (SGS) with a fixed mean, standard deviation, and variogram ranges. The porosity fields which are also generated using SGS have a variable variogram structure and azimuth. A cloud transform of the porosity field is used to generate the permeability and initial saturation fields. After appropriate training, the AI agent is expected to provide an optimized field development plan for an optimization scenario randomly sampled from this defined range of parameters. The generalization and applicability of the resulting AI agent will depend on the optimization problem parameters and their distributions. The number of learnable parameters of the neural network and computational complexity of the training process of the AI increases as more parameters and larger distributions are considered.

\vspace{10pt}

 \begin{table}[!htb]
    \centering
    \caption{Distribution of parameters for 2D and 3D oil-water system field development optimization. $U[a,b]$ denotes uniform distribution over the range of $[a,b]$. $\{a_1,a_2,\cdots,a_n\}$ indicate uniform probability distribution over the $n$ discrete options.}
     \begin{tabular}{cccc}
		\hline
        Number & Variable & Symbol & Distribution  \\
        \hline
        1 &  Grid size in x-direction (ft)  & $dx$ & U[500, 700] \\ 
        2 &  Grid size in y-direction (ft)  & $dy$ & U[500, 700] \\
        3 &  Grid thickness (ft)            & $dz$ & SGS(1000,100,30,60) \\
        4 &  Variogram azimuth ($^0$)            & $ang$ & U[0, 90] \\
        5 &  Variogram structure            & $struct$ & \{Gaussian, exponential\} \\
        6 &  Porosity            & $\phi$ & SGS(U[0.15,0.25],U[0.01,0.07],2,8) \\
        7 & Permeability (md) &  $k$  & cloud transform from $\phi$ \\ 
        8 & Vertical to horizontal permeability ratio (3D) &  $kvkh$  & logUnif[0.001, 0.1] \\ 
        9 &  Active cell indicator & $active$   & Random elliptical \\
        10 &  Datum depth (ft)            & $d_{datum}$ & U[5000, 36000] \\
        11 &  Depth from datum (ft)            & $D$ & SGS(1000,1000,30,60) \\
        12 &  Pressure gradient (psi/ft)            & $p_{grad}$ & U[0.7, 1] \\
        13 & Initial water saturation &  $S_{winit}$  & cloud transform from $\phi$ \\ 
        14 &  Oil reference formation volume factor            & $B_{o,ref}$ & U[1, 1.5] \\
        15 &  Water reference formation volume factor            & $B_{w,ref}$ & U[1, 1.5] \\
        16 &  Oil compressibility (psi$^{-1}$)          & $c_o$ & U[1e-6, 4e-6] \\
        17 &  Water compressibility (psi$^{-1}$)            & $c_w$ & U[2e-6, 5e-6] \\
        18 &  Oil specific gravity           & $\gamma_o$ & U[0.8, 1] \\
        19 &  Water density (lbs/ft$^3$)            & $\rho_w$ & U[62, 68] \\
        20 &  Oil reference viscosity (cp)           & $\mu_{o,ref}$ & U[2, 15] \\
        21 &  Water reference viscosity (cp)            & $\mu_{w,ref}$ & U[0.5, 1] \\
        22 &  Residual oil saturation            & $S_{or}$ & U[0.05, 0.2] \\
        23 &  Connate water saturation            & $S_{wc}$ & U[0.05, 0.2] \\
        24 &  Oil relative permeability endpoint            & $k_{roe}$ & U[0.75, 0.95] \\
        25 &  Water relative permeability endpoint            & $k_{rwe}$ & U[0.75, 0.95] \\
        26 &  Corey oil exponent           & $n_{o}$ & U[2, 4] \\
        27 &  Corey water exponent           & $n_{w}$ & U[2, 4] \\
        28 &  Producer BHP (\% pressure at datum)          & $p^w$ & U[0.3, 0.7] \\
        29 &  Producer skin          & $s$ & U[0, 2] \\
        30 &  Oil price per bbl (\$)          & $p_o$ & U[40, 60] \\
        31 &  Water production cost (\% oil price)          & $c_{wp}$ & U[0, 0.02] \\
        32 &  Operating cost per bbl (\$)          & $c_{opex}$ & U[8, 15] \\
        33 &  Drilling cost per well (\$)          & $c_{capex}$ & U[1e8, 5e8] \\
        34 &  Discount rate (\%)          & $b$ & U[9, 11] \\
        35 &  Drilling time per well (days)          & $d_{time}$ & \{90, 120, 180, 240\} \\
        36 &  Project duration (years)          & $p_{time}$ & U[15, 20] \\
        37 &  Watercut limit (\%)          & $wc_{limit}$ & U[60, 98] \\
        38 &  Max. well liquid production rate (bbl/day)          & $q_{l,max}$ & U[1e4, 2.5e4] \\
        
        \hline
     \end{tabular}
     \label{tab:range_of_param}
\end{table}

As pointed out in~\citet{He2021Deep}, the parameters listed in Table~\ref{tab:range_of_param} do no affect the model independently. Using them directly as input to the deep neural network increase the complexity and the ill-conditioning of the problem. Therefore, we define the state of the environment using parameter groups that aid in reducing the number of input channels provided to the artificial intelligence agent. These input channels include parameter groups that appear in the discretized form of the oil-water immiscible flow problem (Eq.~\ref{Eqn:discretized_equation}) and the reward. Additional input channels are also included to define the operational constraints imposed on the field development. The constraints considered include the drilling duration per well, watercut limit and the maximum well liquid production rate. The list of input channels used to define the state of the environment are given in Table~\ref{tab:env_obs}. 

There are 17 input channels considered for the 2D subsurface system while 18 (including the z-directional tramsmissibility) are used for the 3D system. The observation at any given drilling stage ($o_t$) is represented by $o_t \in \mathbb{R}^{N_x \times N_y \times N_c}$, where $N_c$ defines the number of channels. For the 2D case $N_c = 17$, where each scalar property (such as the drilling time or water cut constraint) is converted to a 2D map of constant value. Out of the 18 input channels for the 3D system, 10 (e.g. pressure, saturation, well index, e.t.c) vary spatially and thus the values for each layer of the reservoir model is included in the input channel. This means in the 3D case, $N_c = 10N_z + 8$, where $N_z$ is the number of layers in the 3D reservoir model.

Once an optimization scenario is defined during training, the static input channels are computed and saved. After advancing the flow simulation to any given drilling stage, the dynamic properties are extracted and stacked with the static properties. This stacked input channels are then provided to the agent as the state of the environment at that given drilling stage. In order to improve the efficiency of the training process, all the input channels (except for the well location mask that is zero everywhere and 1 only at regions where wells are located) are normalized to be very close to 0 to 1 range. Due to the highly skewed distribution of the transmissibilities and well index, they are scaled with a nonlinear scaling function as done in \cite{He2021Deep}. Other input channels are scaled using the linear min-max scaling function.   

\vspace{10pt}

 \begin{table}[!htb]
    \centering
    \caption{List of the input channels used to define the state for 2D and 3D oil-water system field development optimization}
     \begin{tabular}{ccc}
		\hline
        Number & Input Channel & Static/Dynamic  \\
        \hline
        1 &  Pressure                           & Dynamic  \\ 
        2 &  Water saturation                   & Dynamic \\
        3 &  x-directional transmissibility     & Static \\
        4 &  y-directional transmissibility     & Static \\
        5 &  z-directional transmissibility (3D) & Static \\
        6 &  Oil Accumulation                   & Dynamic \\
        7 &  Water mobility                     & Dynamic \\
        8 &  Oil mobility                       & Dynamic \\
        9 &  Well index                         & Static \\
        10 &  Producer drawdown                 & Dynamic \\
        11 &  Well location mask                & Dynamic \\
        12 &  Well cost to net oil price ratio  & Static \\
        13 &  Water production cost             & Static \\
        14 &  Current discount rate             & Dynamic \\
        15 &  Remaining production time         & Dynamic \\
        16 &  Max. liquid production rate       & Static \\
        17 &  Drilling time                     & Static \\
        18 &  Water cut constraint              & Static \\
        
        \hline
     \end{tabular}
     \label{tab:env_obs}
\end{table}

\subsection{Proximal policy optimization}

Our goal is to determine the learnable parameters (defined by the weights of a neural network) of the policy function that maximizes the expected cumulative reward as posed in the optimization problem given in Eq.~\ref{Eqn:policy_opt}. In this work, we use a policy gradient method in which the expected cumulative reward is maximized by performing gradient descent on the parameters of the policy. The vanilla policy gradient method is susceptible to performance deterioration due to the instability that may be introduced by the update step during gradient descent. Several variants of the policy gradient method have been proposed to improve the stability of the optimization. Some variants of the policy gradient method include the trust region policy optimization (TRPO)~\citep{schulman2015trust} and proximal policy optimization (PPO)~\citep{schulman2017proximal}.

The PPO algorithm proposes a surrogate objective that improves the policy optimization by guaranteeing monotonic policy improvement. Following \cite{He2021Deep}, we now briefly describe the implementation of PPO used in this work.

The PPO loss or objective function given in Eq.~\ref{Eqn:total_loss} contains four components: the surrogate policy loss ($L^{\pi}$), the Kullback–Leibler (KL) divergence penalty ($L^{kl}$), the value function loss ($L^{vf}$), and the entropy penalty ($L^{ent}$).

\begin{equation}\label{Eqn:total_loss}
    L^{PPO} = L^{\pi}+c_{kl}L^{kl}+c_{vf}L^{vf}+c_{ent}L^{ent}
\end{equation}

\noindent 
where $c_{kl}$, $c_{vf}$ and $c_{ent}$ are user-defined weighting factors for the KL divergence, value function and entropy terms, respectively.

The surrogate policy loss ($L^{\pi}$) that directly maximizes the expected cumulative reward is given by:

\begin{equation}\label{Eqn:policy_loss_ppo}
    L^{\pi} = \mathbb{E}_t[\min(r_t(\theta) A_t,\textit{clip}(r_t(\theta), 1-\epsilon, 1+\epsilon) A_t)]
\end{equation}
where $r_t(\theta)$ quantifies the policy change and it is defined as the ratio of the old policy $\pi(a_t | s_t, \theta_{old})$ to the new policy $\pi(a_t | s_t, \theta)$. The term $\textit{clip}(r_t(\theta), 1-\epsilon, 1+\epsilon)$ removes the incentive to change the policy beyond a pre-defined step size limit $\epsilon$. This prevents large updates of the parameters that could lead to deterioration of the policy. The advantage function $A_t$ defines how good an action is for a specific state relative to a baseline. Following the generalized advantage estimation (GAE) framework proposed by \cite{schulman2015high}, $A_t$ is defined by:

\begin{equation}
    A_t^{\textit{GAE}(\gamma,\lambda)}(t) = \sum_{l=t}^{T}\left(\gamma\lambda\right)^{l-t}\left(r_l+\gamma V^\pi(s_{l+1})-V^\pi(s_{l}) \right)
    \label{Eqn:advantage_gae}
\end{equation}
where $\gamma$ and $\lambda$ are hyperparameters that control the bias and variance introduced by the various terms in the summation. The value function $V^\pi(s_t)$, given in Eq.~\ref{Eqn:value_func}, represents the expected total reward of being in state $s_t$ and then following the current policy $\pi$. The value function is parameterized by the weights of a neural network $\psi$.
\begin{equation}\label{Eqn:value_func}
    V^\pi(s_{t}) = \mathbb{E}\left[\sum^T_{l=t} r_l\right]
\end{equation}

The value function is learned by minimizing the value-function loss $L^{vf}$ which is given by:

\begin{equation}\label{Eqn:value_function_loss}
    L^{\textit{vf}} = \mathbb{E}_t\left[\max\left(\left(V_\psi(s_{t})-V_{target}(s_{t})\right)^2,\left(V_{\psi_{old}}+\textit{clip}(V_\psi(s_{t})-V_{\psi_{old}}(s_{t}),-\eta,\eta)-V_{target}(s_t)\right)^2\right)\right],
\end{equation}
where the $V_{target}(s_{t})$ is the computed value function from the training samples, $V_{\psi_{old}}$ is the value function using the current parameters of the value function $\psi_{old}$. Equation \ref{Eqn:value_function_loss} essentially tries to minimize the mismatch between the computed and predicted value of the sampled states. The hyperparameter $\eta$ has same effect as $\epsilon$ in Eq.~\ref{Eqn:policy_loss_ppo} and is used to ensure large updates of the value parameters $\psi$ are not allowed.

The KL divergence penalty ($L^{kl}$) serves as an additional term to avoid large policy update and it is given by: 
\begin{equation}\label{Eqn:KL_loss}
    L^{KL} = \mathbb{E}_t\left[D_{\textit{KL}}(\pi(a_t | s_t, \theta)|\pi(a_t | s_t, \theta_{old}))\right],
\end{equation}
where $D_{\textit{KL}}(\pi(a_t | s_t, \theta)|\pi(a_t | s_t, \theta_{old}))$ is the Kullback–Leibler divergence~\citep{kullback1951information} that measures the difference between the old and new policy.

\subsection{Policy and value function representation}

\begin{figure}[htbp]
	\centering
	\includegraphics[width=0.75\textwidth]{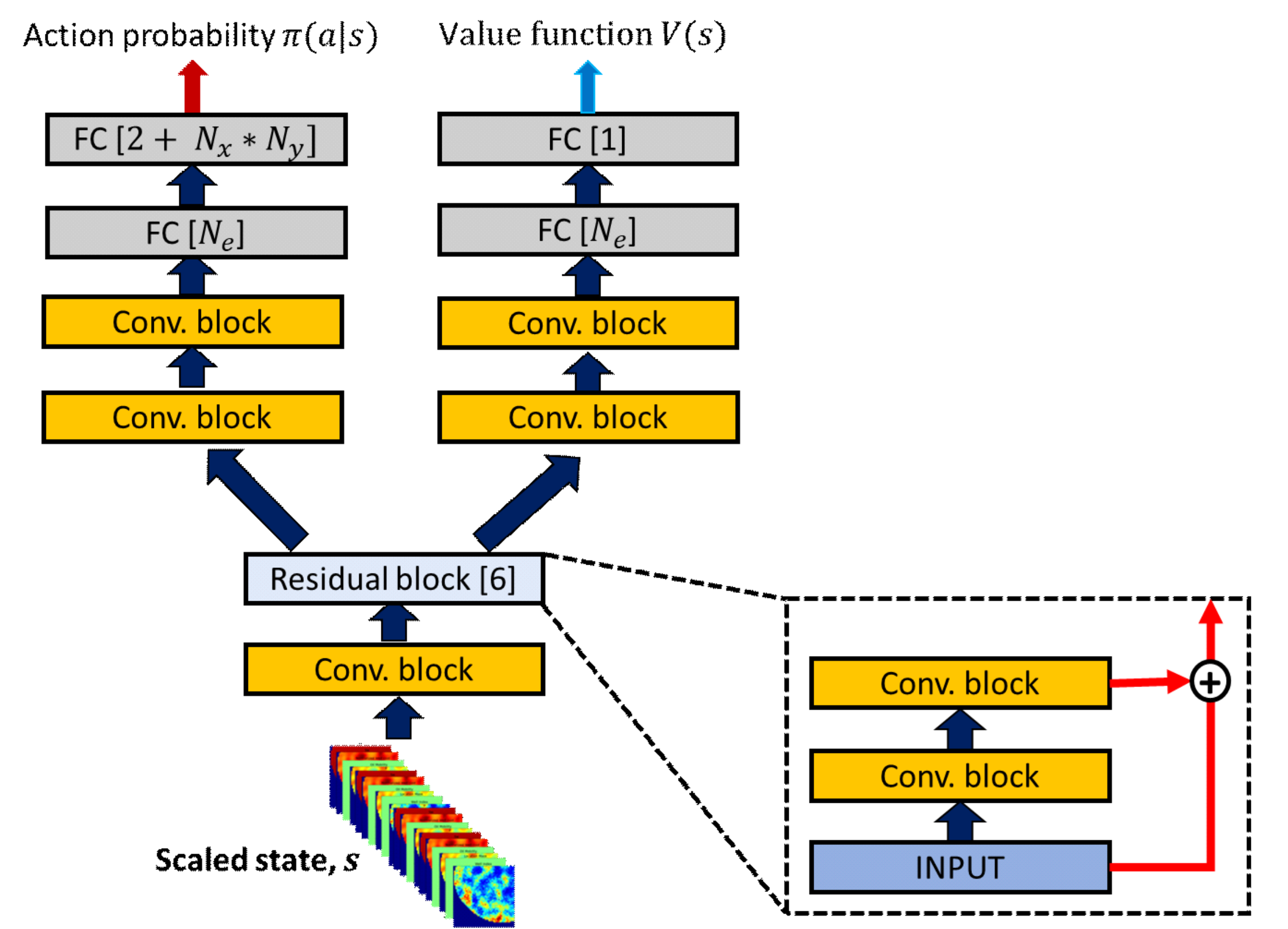}
	\caption{The neural network architecture that defines the policy and value functions. The "conv" block represents a convolutional neural network (CNN) layer followed by rectified linear unit (ReLU). FC[$N_e$] represents a fully connected layer with $N_e$ neurons.}
	\label{fig:cnn}
\end{figure}

We now describe the representation of the policy function $\pi(a|s, \theta)$ and the value function $V^\pi(s_{t})$ with a deep neural network. In our previous work \citep{He2021Deep}, the policy and value networks are represented by different neural networks. Thus the parameters $\theta$ and $\psi$ are independent. In this work, we use the approach in AlphaGo Zero~\citep{silver2017mastering} where the value and policy networks share some layers. In this case, there is an overlap in the parameters $\theta$ and $\psi$. The neural network architecture used in this work is shown in Fig.~\ref{fig:cnn}. The shared layers in the neural network are used to learn features from the state that are relevant to both the policy and the value network. This reduces the computational cost that will otherwise be expensive during training if the policy and value network are represented by different neural networks.

Given the scaled state at any given drilling stage, the shared layers process the state through a series of convolutional operation. The shared layer comprises a "conv" block and six residual blocks~\citep{he2016deep}. The "conv" block is essentially a convolutional neural network (CNN) layer~\citep{krizhevsky2012imagenet} followed by a rectified linear unit (ReLU) activation function~\citep{nair2010rectified}. Interested readers may refer to \cite{He2021Deep} for a brief description of CNN layers and ReLU activation functions. Each residual block contains two "conv" blocks as shown in Fig.~\ref{fig:cnn}. The convolutional operations in the shared layers are performed with 64 kernels of size $3 \times 3$ for the two-dimensional subsurface system. The agent for the three-dimensional system, however, uses 128 kernels for the "conv" blocks in the shared layers. This is primarily because the size of the input channels in the three-dimensional subsurface system is larger than that of the two-dimensional case.

The learned features from the shared layers serves as input to the policy and value arms of the network. These features are further processed with two "conv" blocks in the individual arms. The first and second conv blocks consist of 128 and 2 kernels, respectively, of size $1 \times 1$. The high dimensional output after the convolutional operations are reduced in dimension with an embedding layer~\citep{he2015deep, He2021Deep} which consist of a fully connected (FC) layer with $N_e$ units. Here, $N_e$ is set to 50. The learned embeddings from the policy arm of the network serves as input to an additional fully connected layer which predicts the probability of all actions. The embeddings from the value arm, however, predicts the value of the state which is a scalar quantity.

%===============================================================================
\section{Computational Results}
\label{sec:results}
%===============================================================================

\begin{figure}[!htb]
    \centering
    \begin{subfigure}[b]{0.315\textwidth}
        \includegraphics[width=1\textwidth]{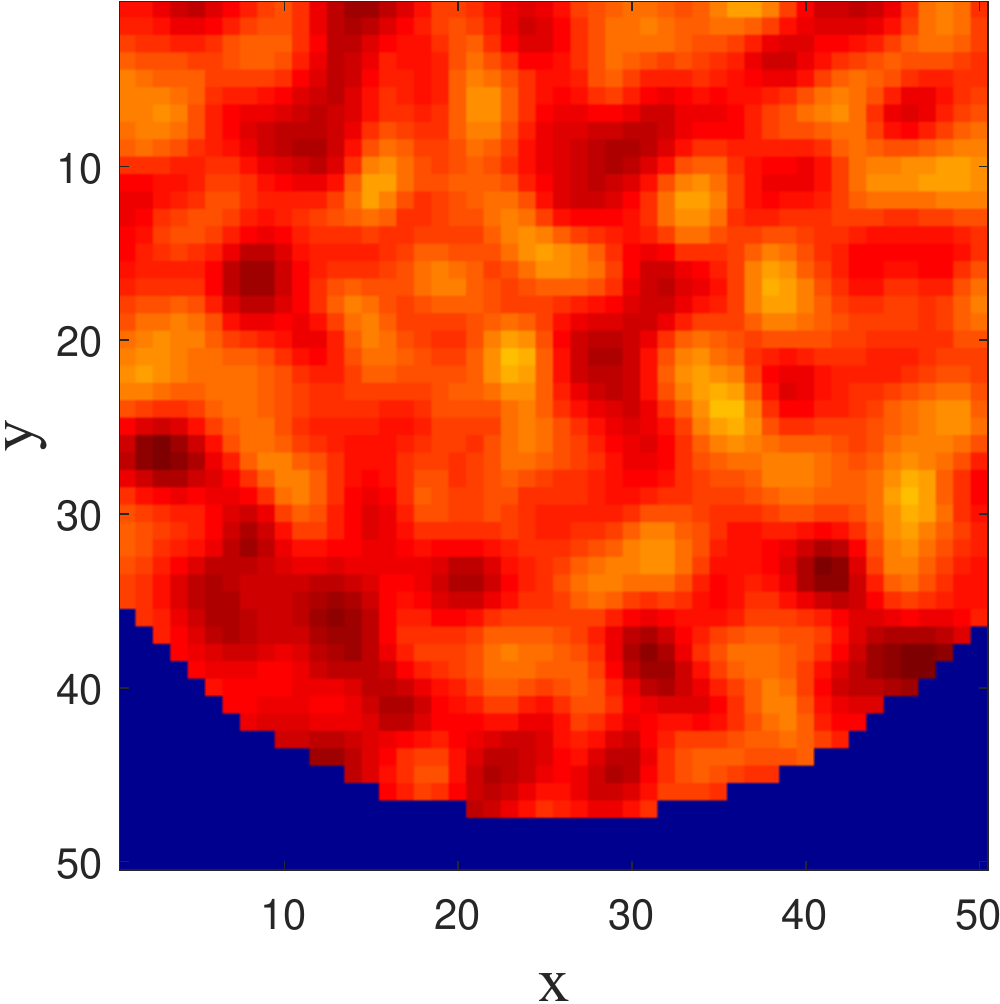}
        \caption{}
    \end{subfigure}%
    ~
    \begin{subfigure}[b]{0.315\textwidth}
        \includegraphics[width=1\textwidth]{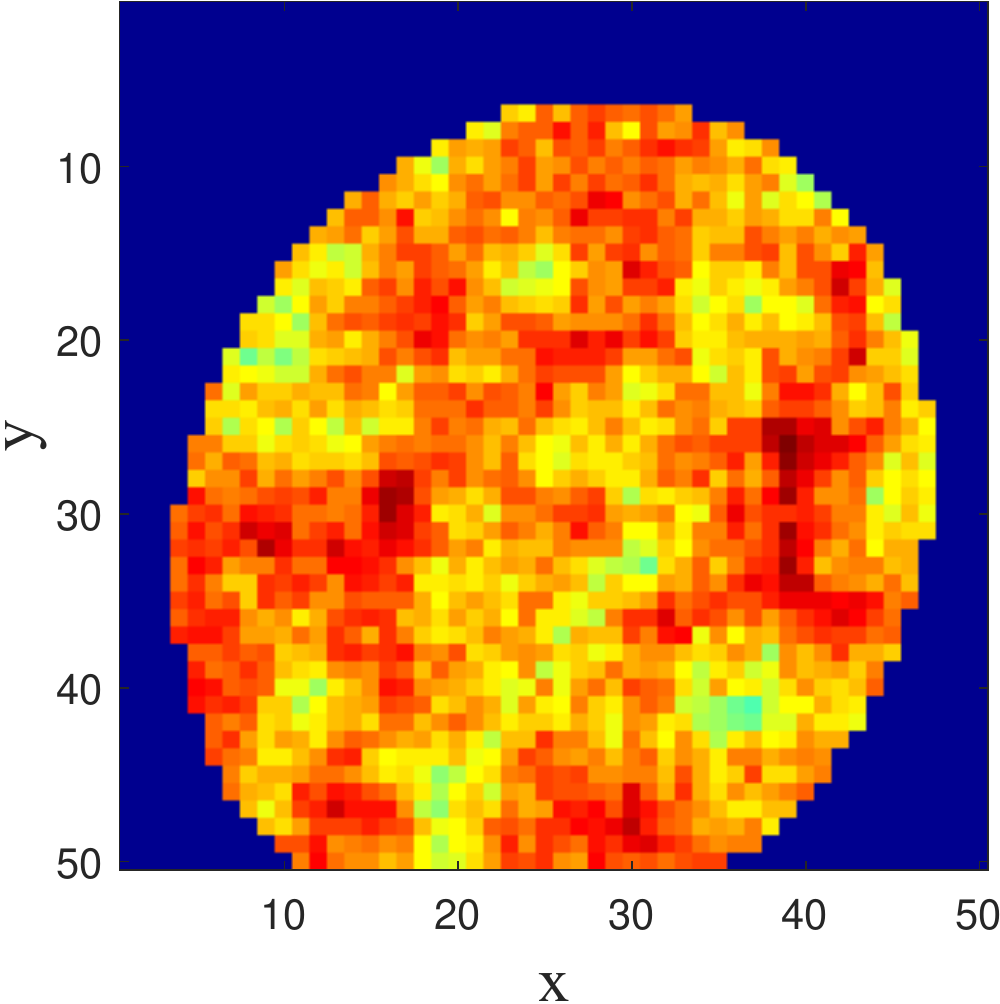}
        \caption{}
    \end{subfigure}%
    ~
    \begin{subfigure}[b]{0.365\textwidth}
        \includegraphics[width=1\textwidth]{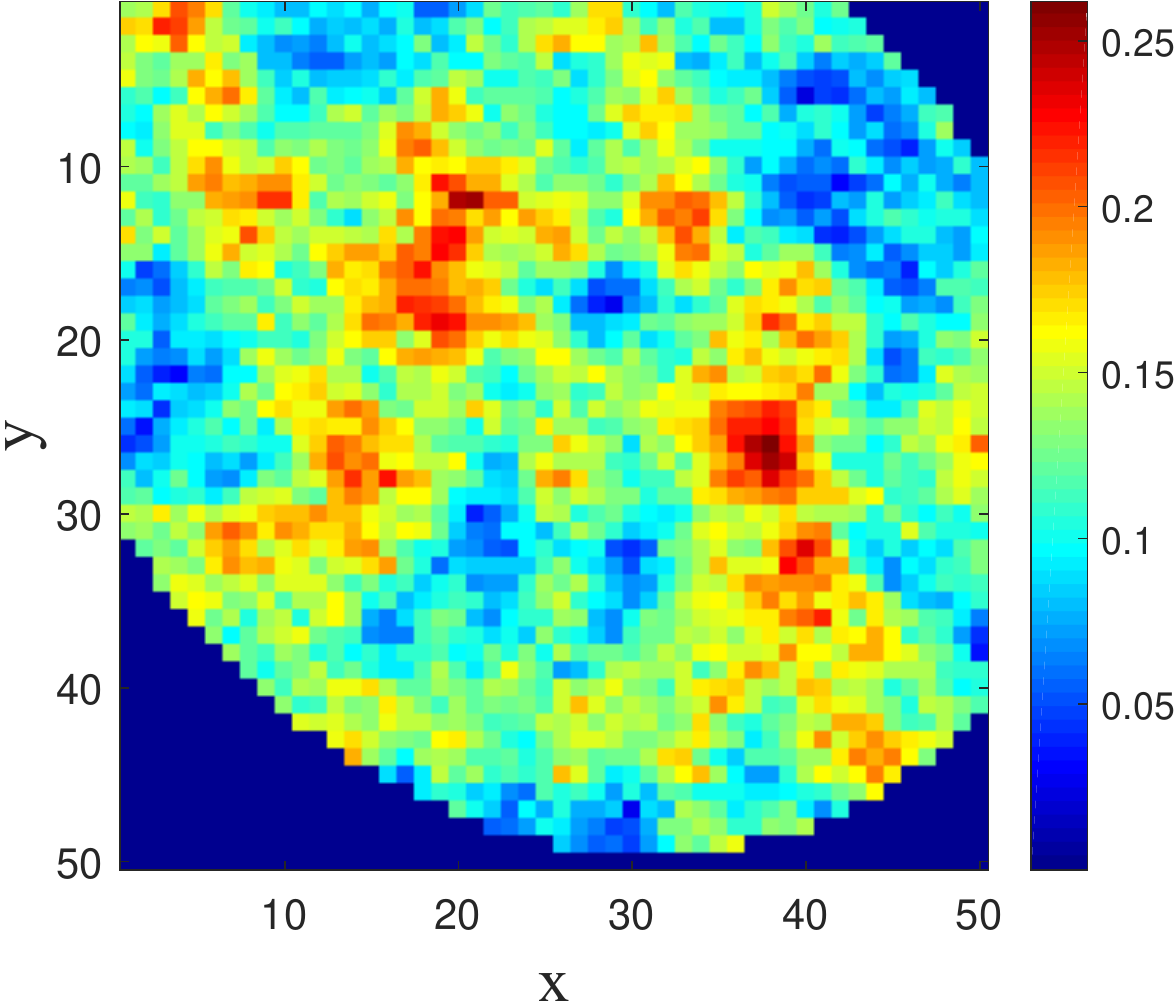}
        \caption{}
    \end{subfigure}%
    
    \begin{subfigure}[b]{0.34\textwidth}
        \includegraphics[width=1\textwidth]{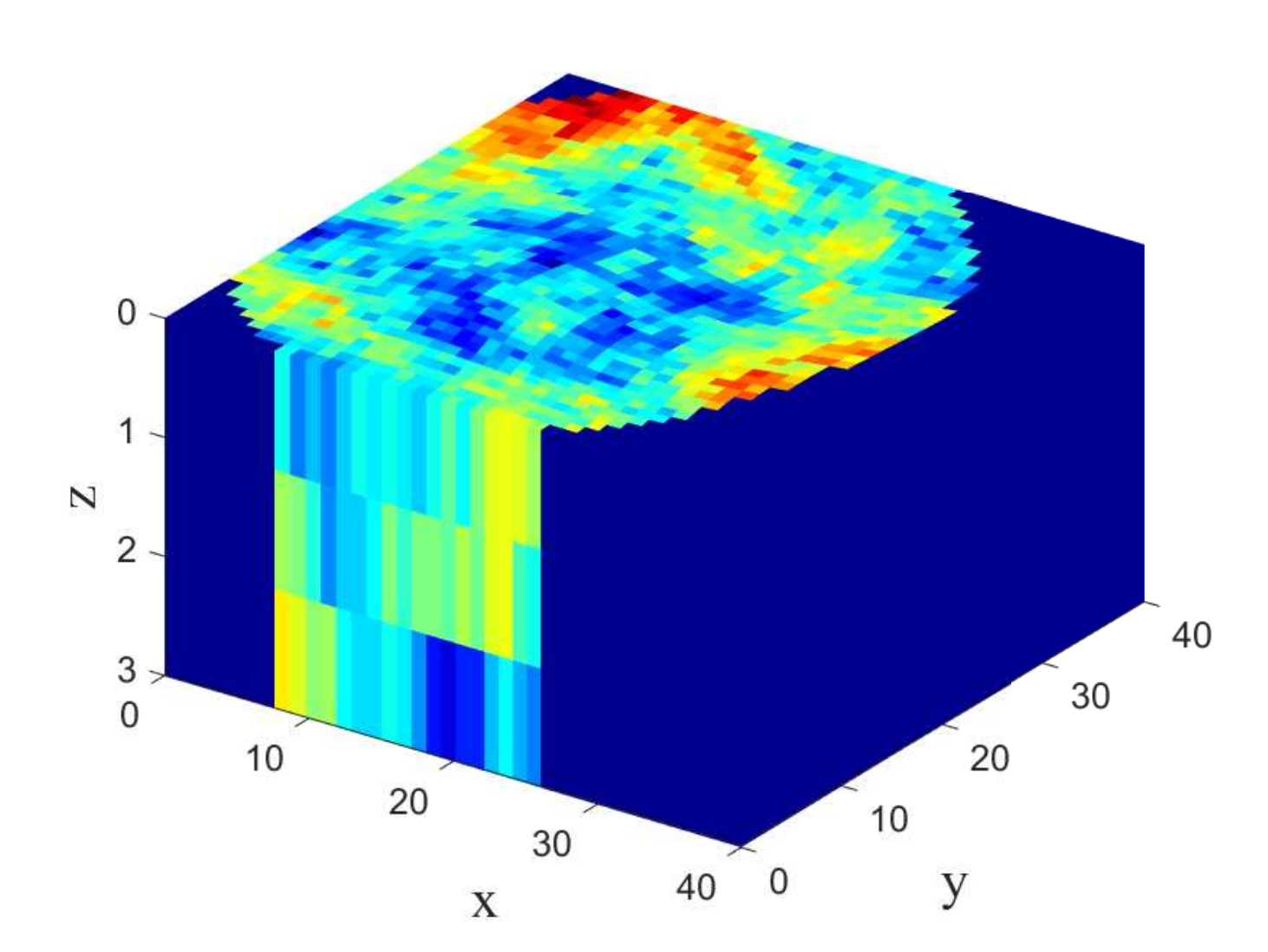}
        \caption{}
    \end{subfigure}%
    ~
    \begin{subfigure}[b]{0.34\textwidth}
        \includegraphics[width=1\textwidth]{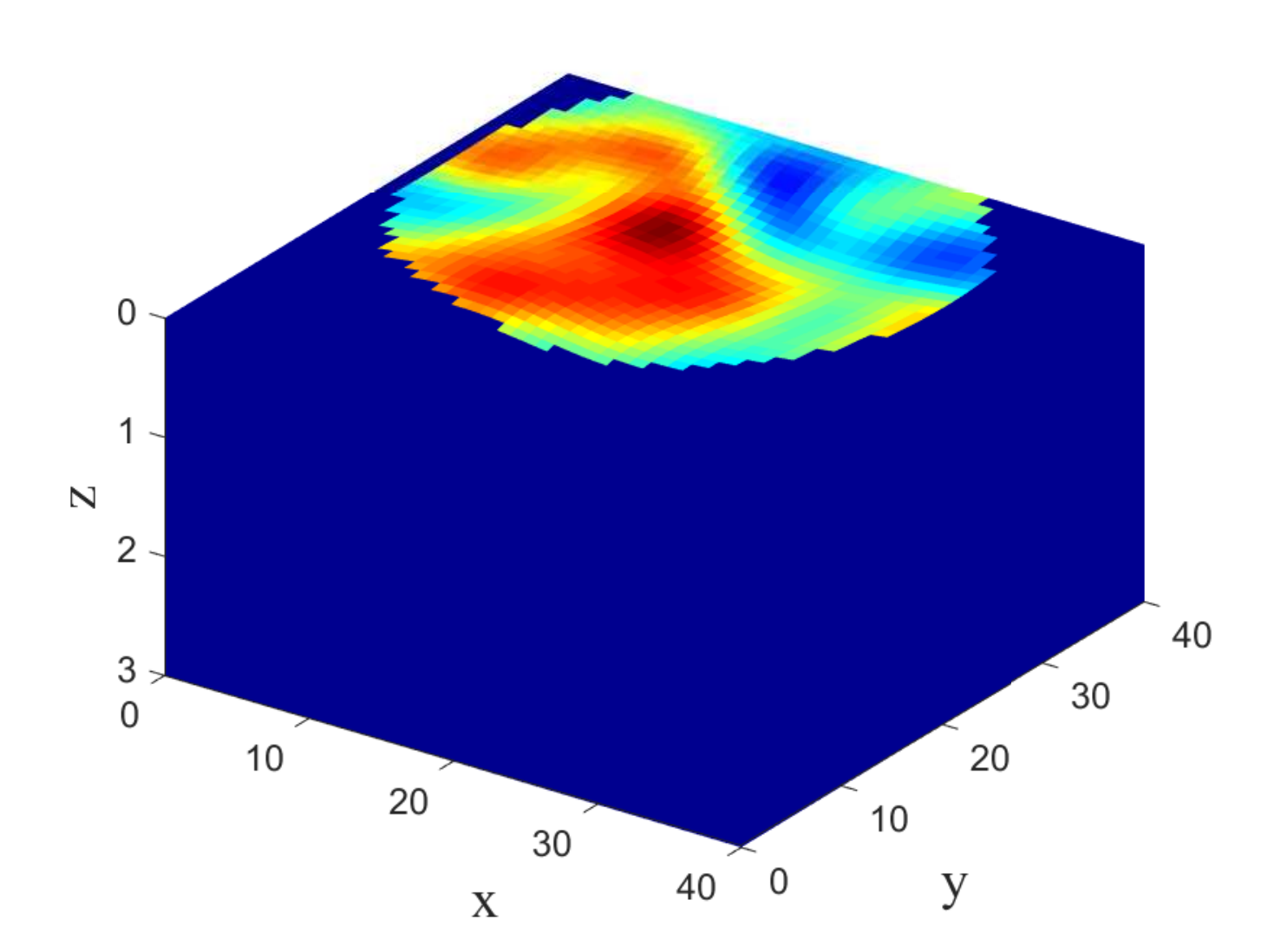}
        \caption{}
    \end{subfigure}%
    ~
    \begin{subfigure}[b]{0.35\textwidth}
        \includegraphics[width=1\textwidth]{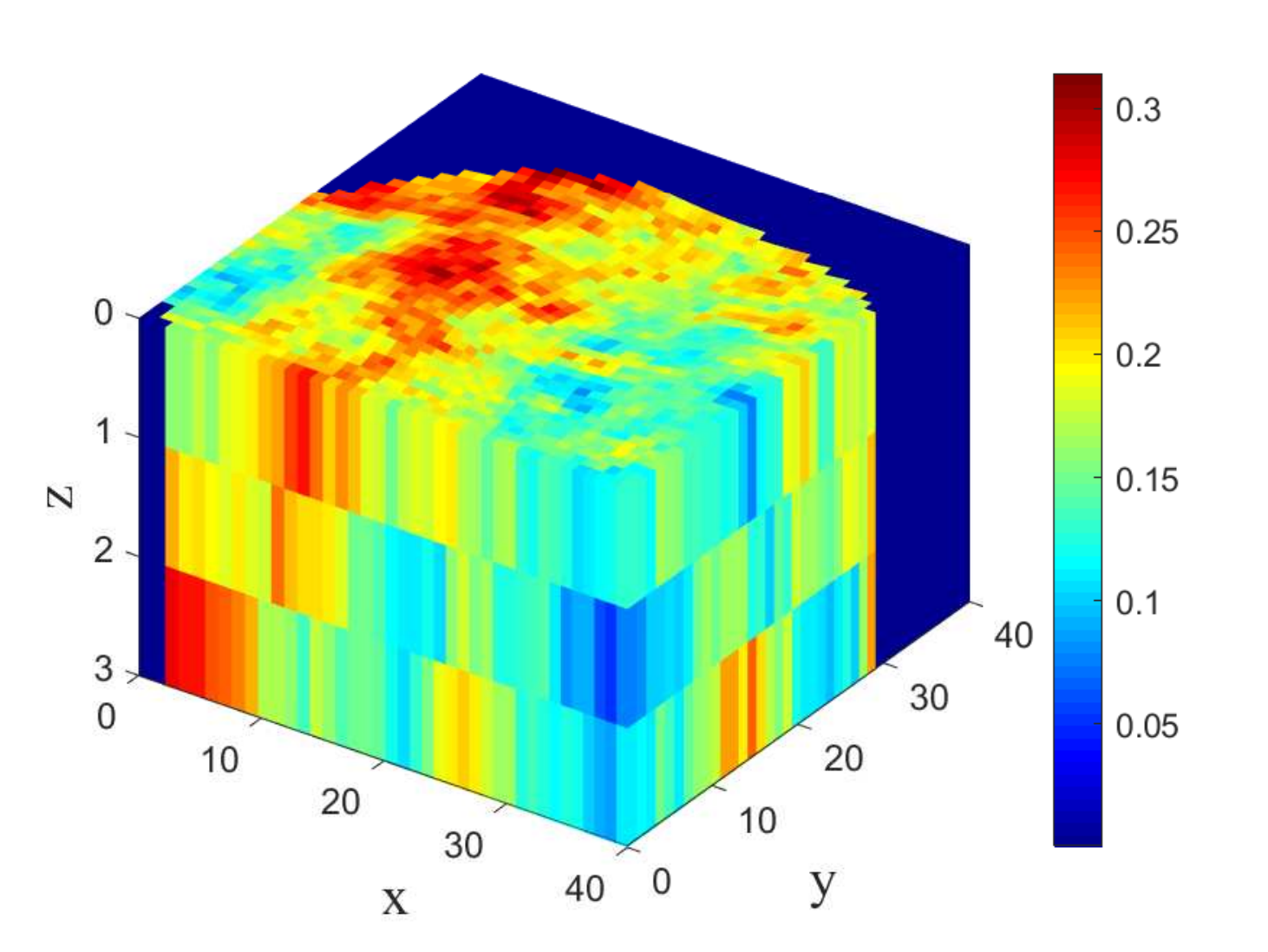}
        \caption{}
    \end{subfigure}%

    \caption{Porosity fields for random training scenarios of the 2D and 3D models sampled from the parameter distributions given in Table~\ref{tab:range_of_param}.}
    \label{fig:training_scenarios}
\end{figure}

In this section, we evaluate the performance of the artificial intelligence agents for field development optimization in two- and three-dimensional subsurface systems. The 2D reservoir model is of dimension $50 \times 50$, while that of the 3D case is $40 \times 40 \times 3$. A maximum of 20 production wells are considered for the field development with one well drilled per drilling stage. The number of drilling stages for each specific field development scenario depends on the total production period and drilling duration sampled from Table~\ref{tab:range_of_param}. Wells operate at the sampled bottom-hole pressures unless a maximum (sampled) water cut is reached, at which point the well is shut in. Porosity fields for three random training scenarios of the 2D and 3D models are shown in Fig.\ref{fig:training_scenarios}~(a)~(b)~(c) and Fig.\ref{fig:training_scenarios}~(d)~(e)~(f), respectively. Note that other variables (such as the constraints, economics and fluid properties) for these training scenarios are, in general, different.

The simulations in both cases are performed using Delft Advanced Research Terra Simulator (DARTS) with operator-based linearization~\citep{khait2017operator, khait2019delft}. The simulator has a python interface that allows for easy coupling with the deep reinforcement learning framework. The light weight nature of the simulator makes it suitable for running millions of simulations with minimal overhead. The overhead is significantly reduced due to the absence of redundant input/output processing. In our implementation, the simulator was further extended to handle dynamic addition of wells between drilling stages. 

The training process utilizes 151 CPU cores for the 2D case and 239 CPU cores for the 3D case. The CPU cores are used to run the simulations in parallel. The training data generated are then passed to 4 GPU cores used for training the deep neural network. The weighting factors for the terms in the loss function are defined based on those proposed in \cite{He2021Deep}. Accordingly, $c_{kl}$, $c_{vf}$ and $c_{ent}$ in Eq.~\ref{Eqn:total_loss} are set to 0.2, 0.1, 0.001. A linear learning rate decay schedule is used for the gradient descent with an initial learning rate of $1e^{-3}$ and a final learning rate of $5e^{-6}$ (at 15 million training samples). It should be noted that each combination of $a_t$, $s_t$ and $r_t$ generated in each drilling stage represents a single training sample. A mini-batch size of 256 and 5 epochs are used for the gradient descent.

For the 2D case, we first compare the performance of the action parameterization used in our previous work \cite{He2021Deep}, with the one proposed in this work. We then benchmark the performance of the 2D artificial intelligence agent with well-pattern drilling agents. The 3D artificial intelligence agent is also compared with the well-pattern drilling agents.

\subsection{Case 1: Two-dimensional system}
\label{sec:2D}

\begin{figure}[!htb]
    \centering
   
    \begin{subfigure}[b]{0.45\textwidth}
        \includegraphics[width=1.1\textwidth]{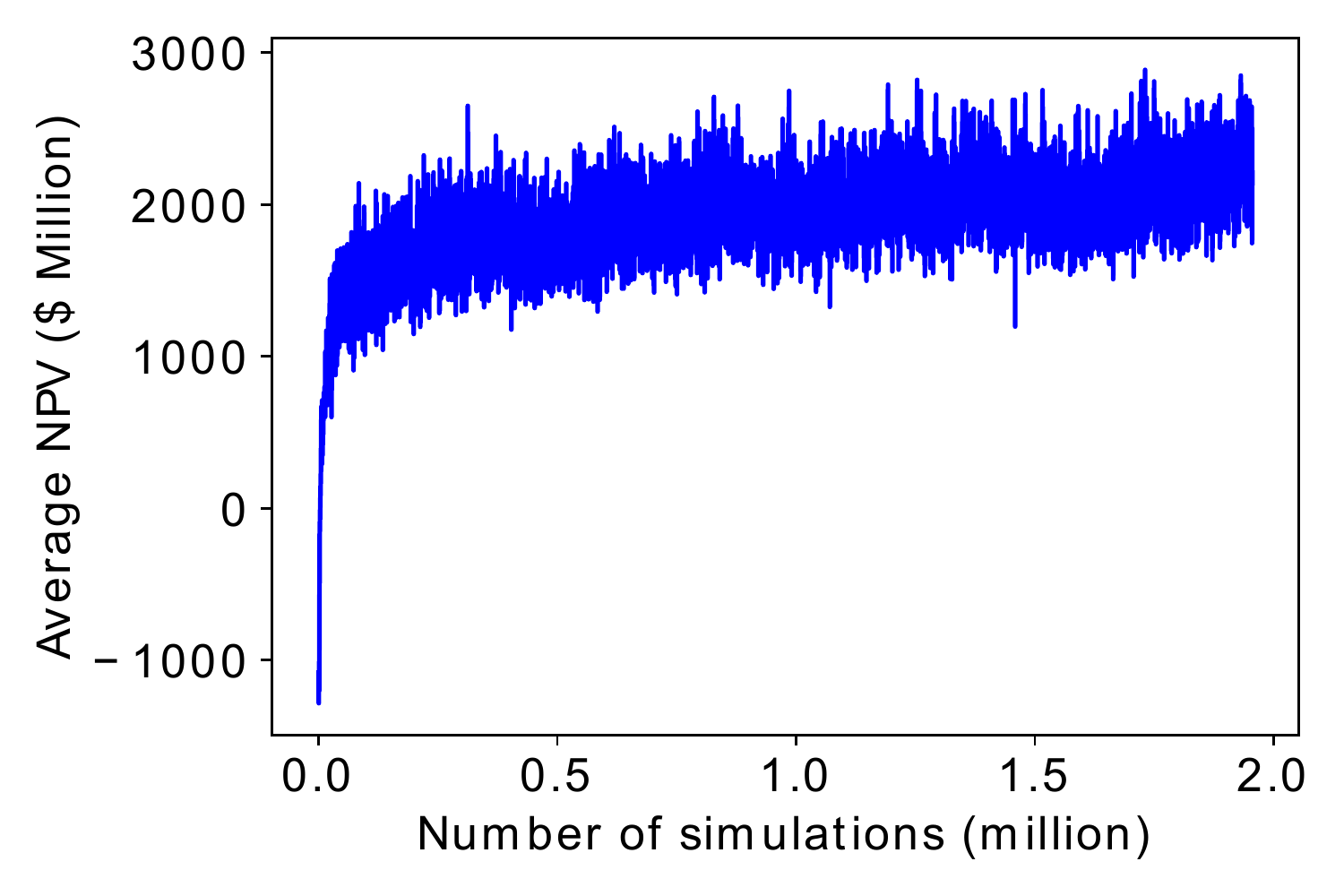}
        \caption{Average NPV}
    \end{subfigure}%
    \hspace{1\baselineskip}
    \begin{subfigure}[b]{0.45\textwidth}
        \includegraphics[width=1.1\textwidth]{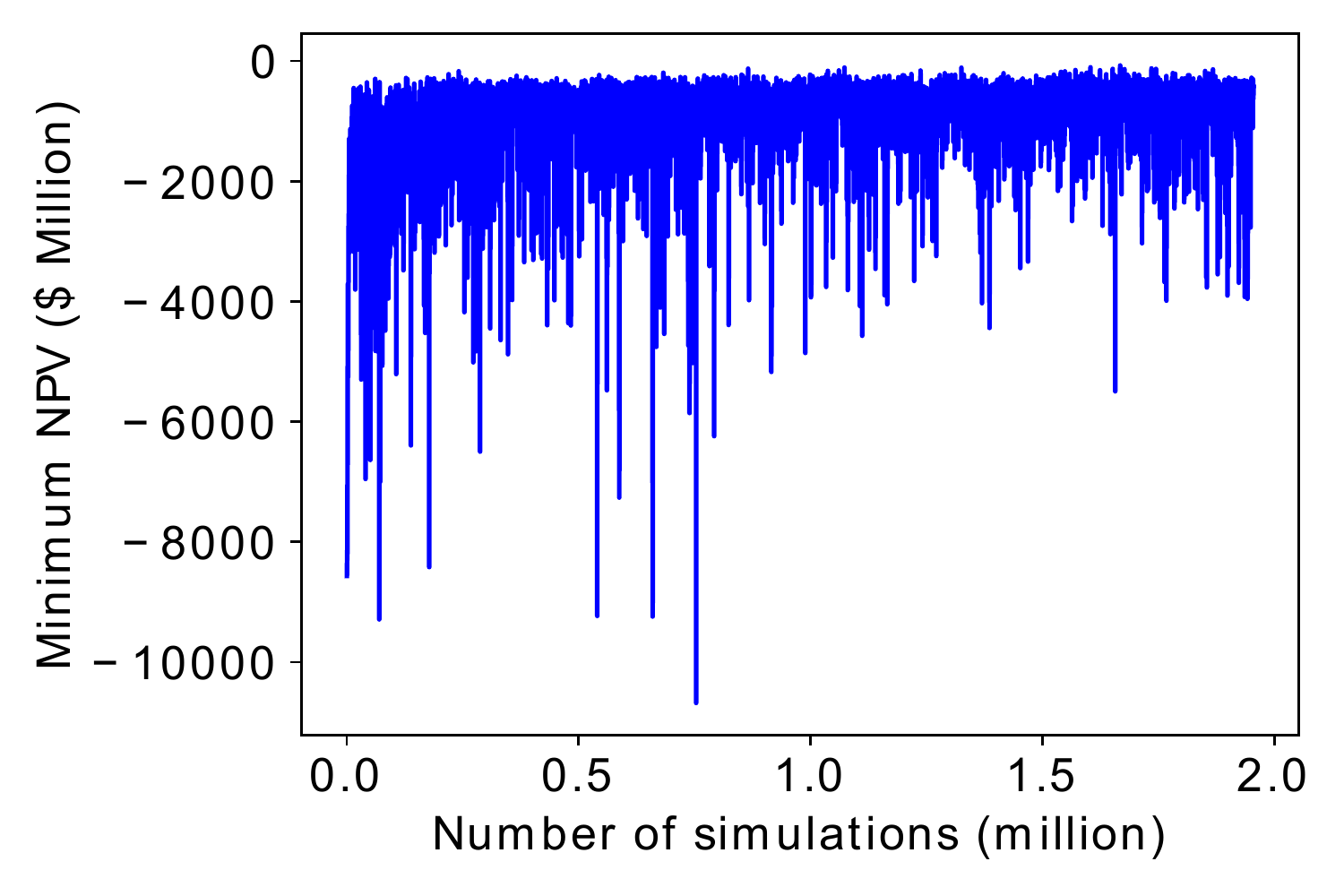}
        \caption{Minimum NPV}
    \end{subfigure}%
    
    \begin{subfigure}[b]{0.45\textwidth}
        \includegraphics[width=1.1\textwidth]{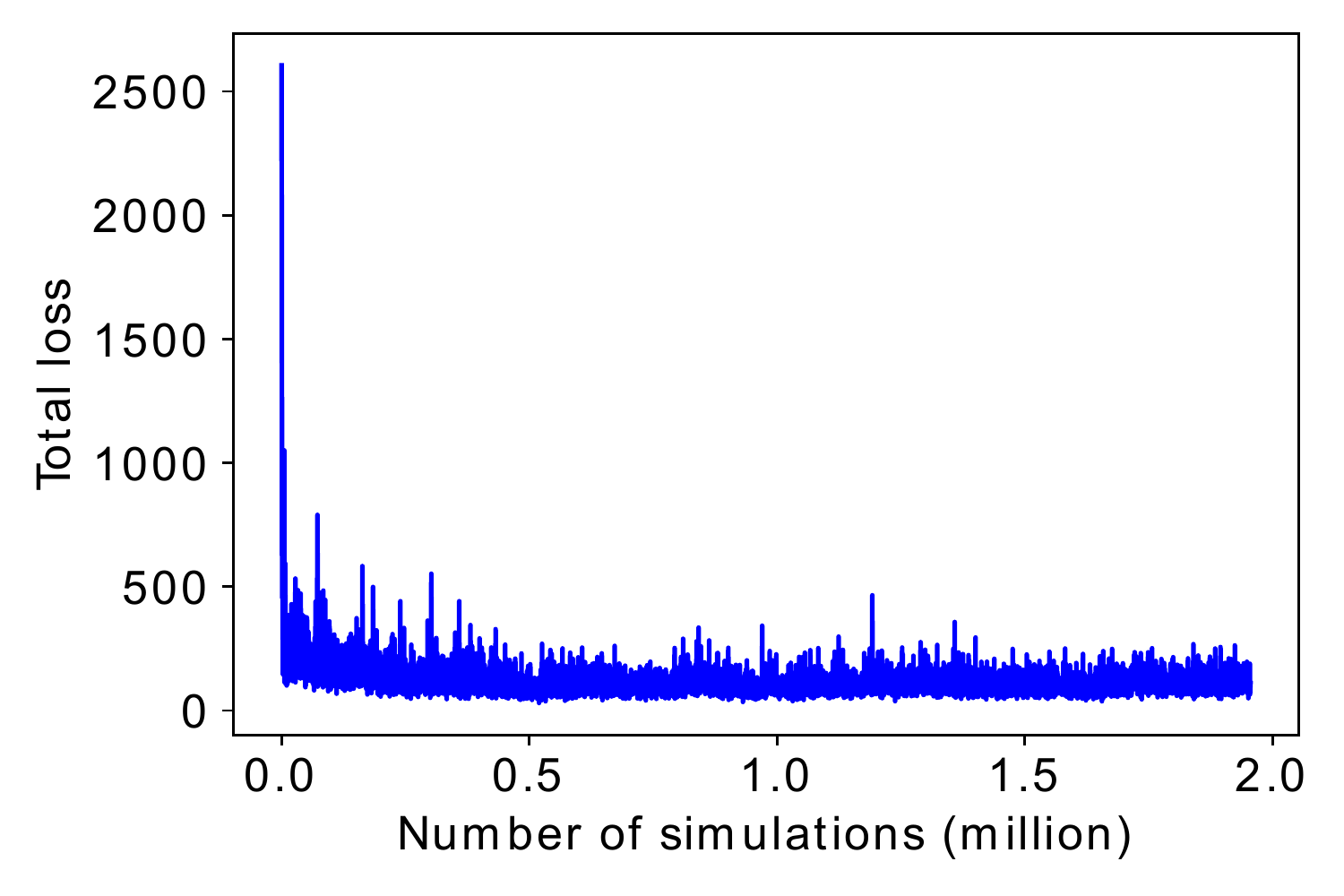}
        \caption{Total loss}
    \end{subfigure}%
    \hspace{1\baselineskip}
    \begin{subfigure}[b]{0.45\textwidth}
        \includegraphics[width=1.1\textwidth]{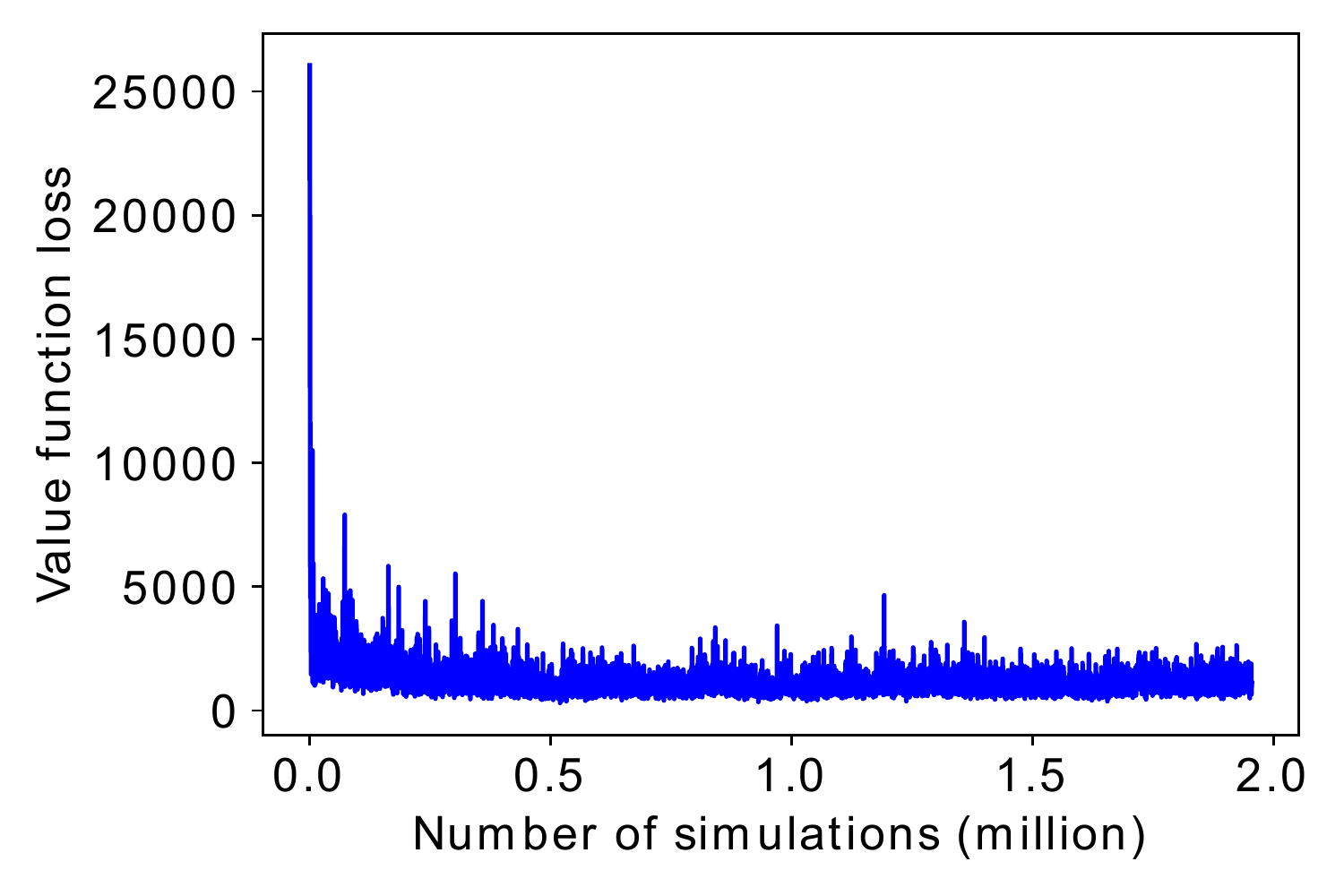}
        \caption{Value function loss}
    \end{subfigure}%
    
    \begin{subfigure}[b]{0.45\textwidth}
        \includegraphics[width=1.1\textwidth]{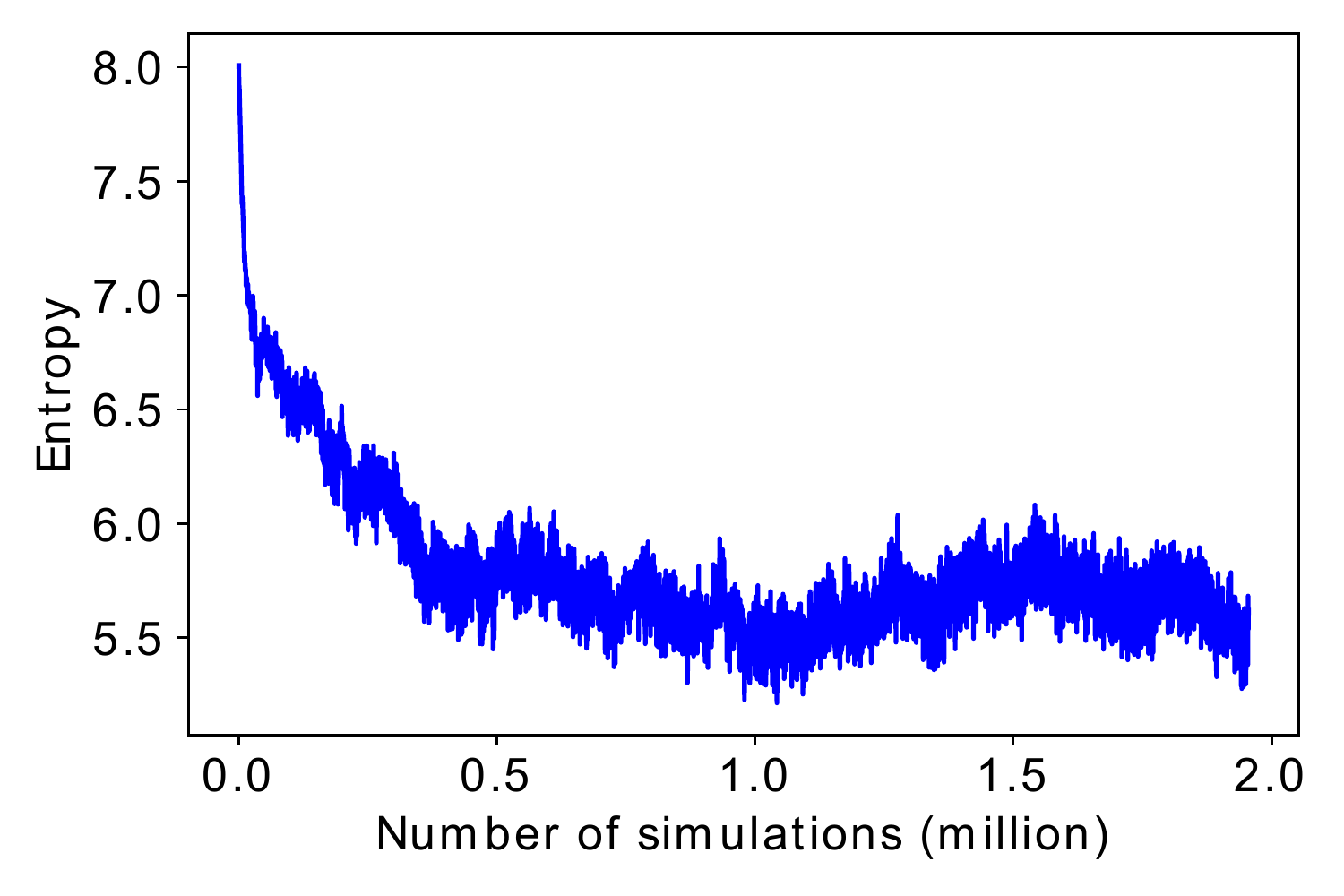}
        \caption{Entropy}
    \end{subfigure}%
    \hspace{1\baselineskip}
    \begin{subfigure}[b]{0.45\textwidth}
        \includegraphics[width=1.1\textwidth]{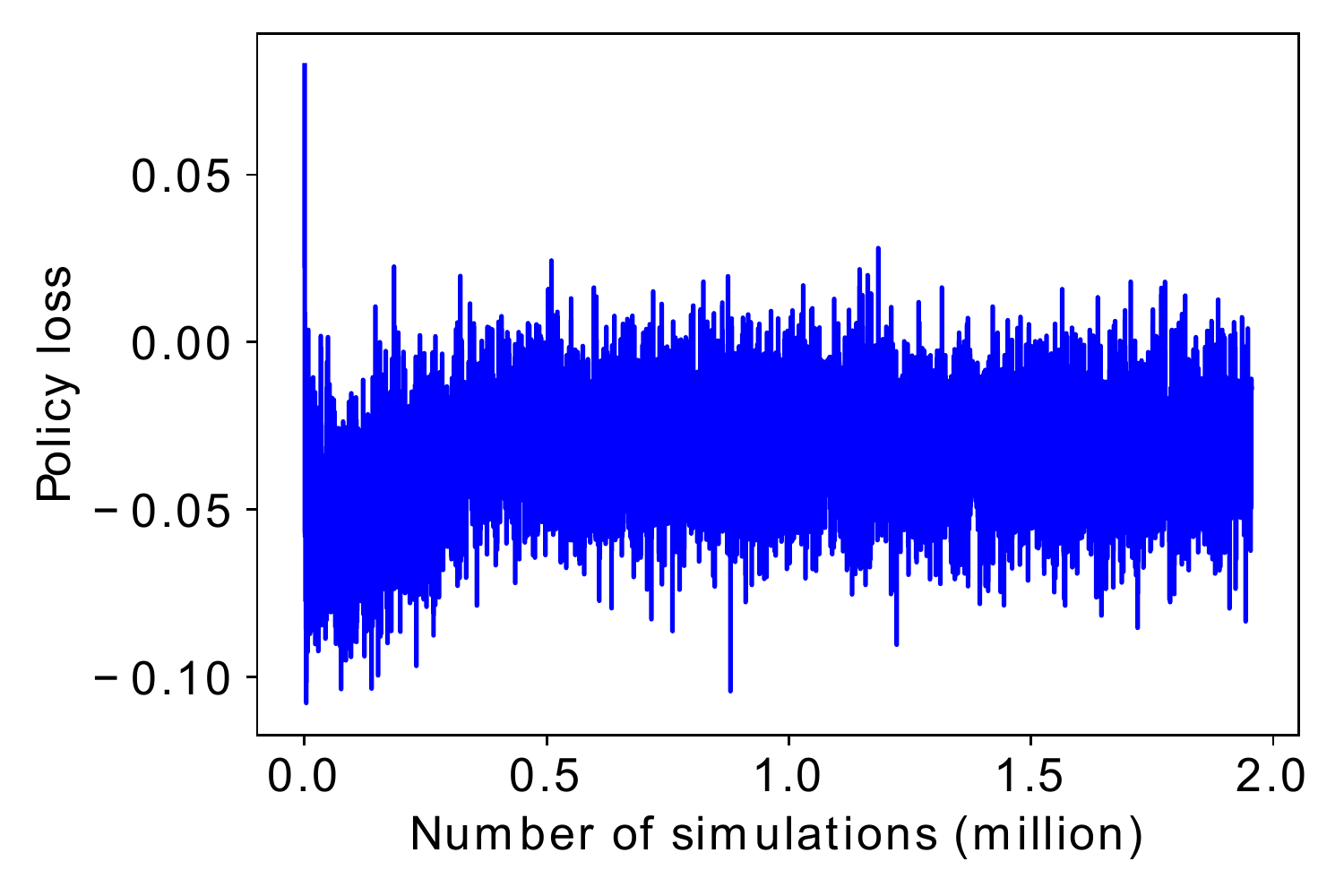}
        \caption{Policy loss}
    \end{subfigure}%
    
     \caption{Evolution of training performance metrics for the dual-action probability distributions for the 2D case.}
	\label{fig:kpi_cur_param}
\end{figure}

We now present results for the 2D case.  Figure~\ref{fig:kpi_cur_param} shows the evolution of some performance indicators for the training process in which the dual-action probability distribution is used. At each of the training iteration, each of the 151 CPUs runs a maximum of two simulations. The set of training data generated (used to train the agent in that specific iteration) are used to compute the performance indicators reported. The training of the agent involves approximately two million simulations, which corresponds to approximately 7,100 equivalent simulations or training iterations.

From Fig.~\ref{fig:kpi_cur_param}~(a), it is evident that the average NPV of the field development scenarios, in general, increases as the training progresses. Starting from a random policy (randomly initiated weights of the deep neural network) which results in negative average NPV, the average NPV increased to more than \$2 billion. The fluctuation in the average NPV is due to the fact that the ease of developing the various field development scenarios, which are randomly generated, varies from iteration to another. 

Figure~\ref{fig:kpi_cur_param}~(b) shows the minimum NPV of the field development scenarios generated at each iteration. Theoretically, the agent should have a minimum NPV of zero as it could choose not to drill any well. However, it should be noted that during the generation of the training data, the action to be taken is sampled from the action distribution of the policy. While this aids in exploring the action space (the entropy loss also encourages exploration) and improves the training performance, it also means the policy is not strictly followed during training. This leads to some fluctuation in the minimum NPV in Fig.~\ref{fig:kpi_cur_param}~(b), but overall the minimum NPV approaches zero. During the application of the agent (after training), the action with the maximum probability is taken. Results for the case in which the policy is followed strictly are presented later.

The PPO total loss, given in Eq.~\ref{Eqn:total_loss}, is shown in Figure~\ref{fig:kpi_cur_param}~(c). The loss, in general, decreases as the optimization progresses. The value function loss, entropy and policy loss are shown in Fig.~\ref{fig:kpi_cur_param}~(d), (e), and (f), respectively. From the figures, we can see that the total loss is dominated by the value function loss. From our limited experimentation, there was no noticeable advantage in the reduction of the weighting factor for the value function loss. The entropy loss which indicates the convergence of the policy can be seen to be decreasing. The policy loss, however, oscillates and does not show any clear trend. This is a common behaviour in reinforcement learning because the definition of the policy loss varies from one iteration to another.

\begin{figure}[!htb]
    \centering
    % INIT
        
        \floatbox[{\capbeside\thisfloatsetup{capbesideposition={left,top},capbesidewidth=7.2cm}}]{figure}[\FBwidth]
{\caption{Evolution of the average NPV using the single-action probability distribution for the 2D case. }\label{fig:prev_act_aver_npv}{\hspace{0.2\baselineskip}}}
    {\includegraphics[width=0.55\textwidth]{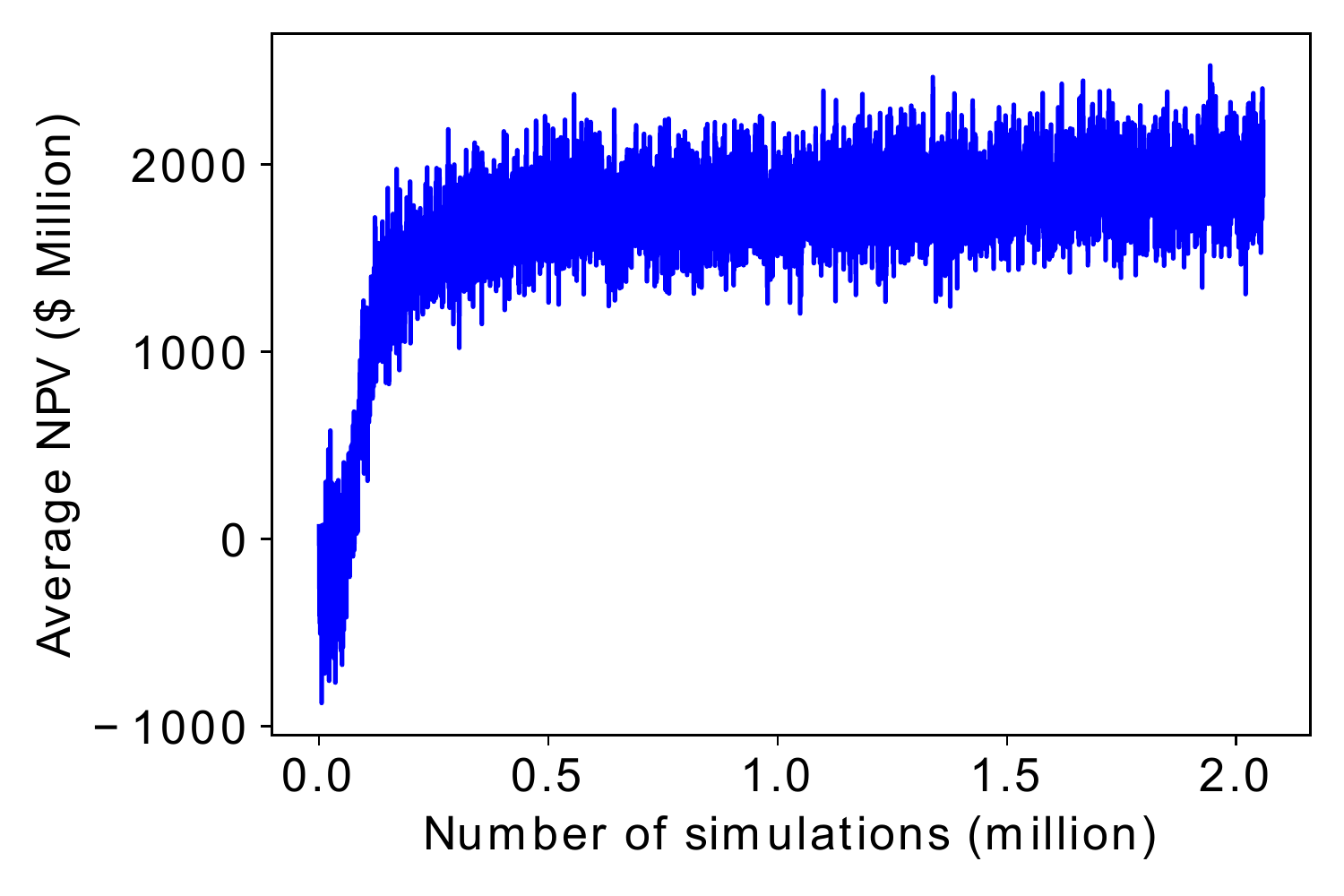}}
\end{figure}

We now compare the performance of the single-action probability distribution used in \cite{He2021Deep} with the dual-action probability distribution. Figure~\ref{fig:prev_act_aver_npv} shows the evolution of the average NPV for the single-action probability distribution. In general, the average NPV increases as the training progresses. However, when compared to Fig.~\ref{fig:kpi_cur_param}~(a), the use of a single-action probability distribution leads to a slow learning in the initial training phase. The result shown here is the best found after several trials. Depending on the initial policy, the learning could be significantly slower than the case shown here. This slow learning is mainly because the learning of the decision not to drill a well is very slow since the action probability distribution is dominated by possible drilling locations. This slow learning increases the computational cost of the training because the timing of the simulation generally increases with number of wells. The timing for the first 100,000 simulations when the single-action probability distribution is used is 11.8 hours. This is in contrast to 7.6 hours for the dual-action probability distribution case, leading to a computational cost saving of approximately 4 hours for the initial 100,000 simulations.

\begin{figure}[!htb]
    \centering
    % INIT
        
        \floatbox[{\capbeside\thisfloatsetup{capbesideposition={left,top},capbesidewidth=7.2cm}}]{figure}[\FBwidth]
{\caption{Evolution of the average NPV for the single and dual-action probability distributions.}\label{fig:comp_prev_cur_fixed}{\hspace{0.2\baselineskip}}}
    {\includegraphics[width=0.55\textwidth]{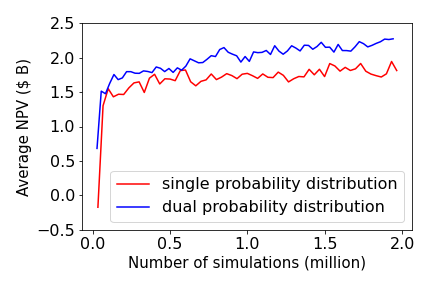}}
\end{figure}

As noted earlier, during training, there is an exploratory aspect to the agent's policy and the field development scenarios are randomly generated. For this reason, we generate 150 random field descriptions, fluid properties, economics and constraints which are used for consistent comparison of the training performance of the single and dual action distributions. Figure~\ref{fig:comp_prev_cur_fixed} shows the evolution of the average NPV for the 150 field development scenarios. The agent after every 100 training iteration is applied for the development of the 150 scenarios. Clearly, the use of dual-action probability distribution outperforms the single-action probability distribution. 

Once the AI agent is trained, it can be used to generate optimized field development plans for any new scenarios within the range of applicability without additional simulations. Its extremely low cost in optimization for new scenarios makes it distinctively different from the traditional optimization methods, which for any new scenario would require hundreds or thousands of simulations.

\begin{figure}[!htb]
    \centering
   
    \begin{subfigure}[b]{0.43\textwidth}
        \includegraphics[width=1\textwidth]{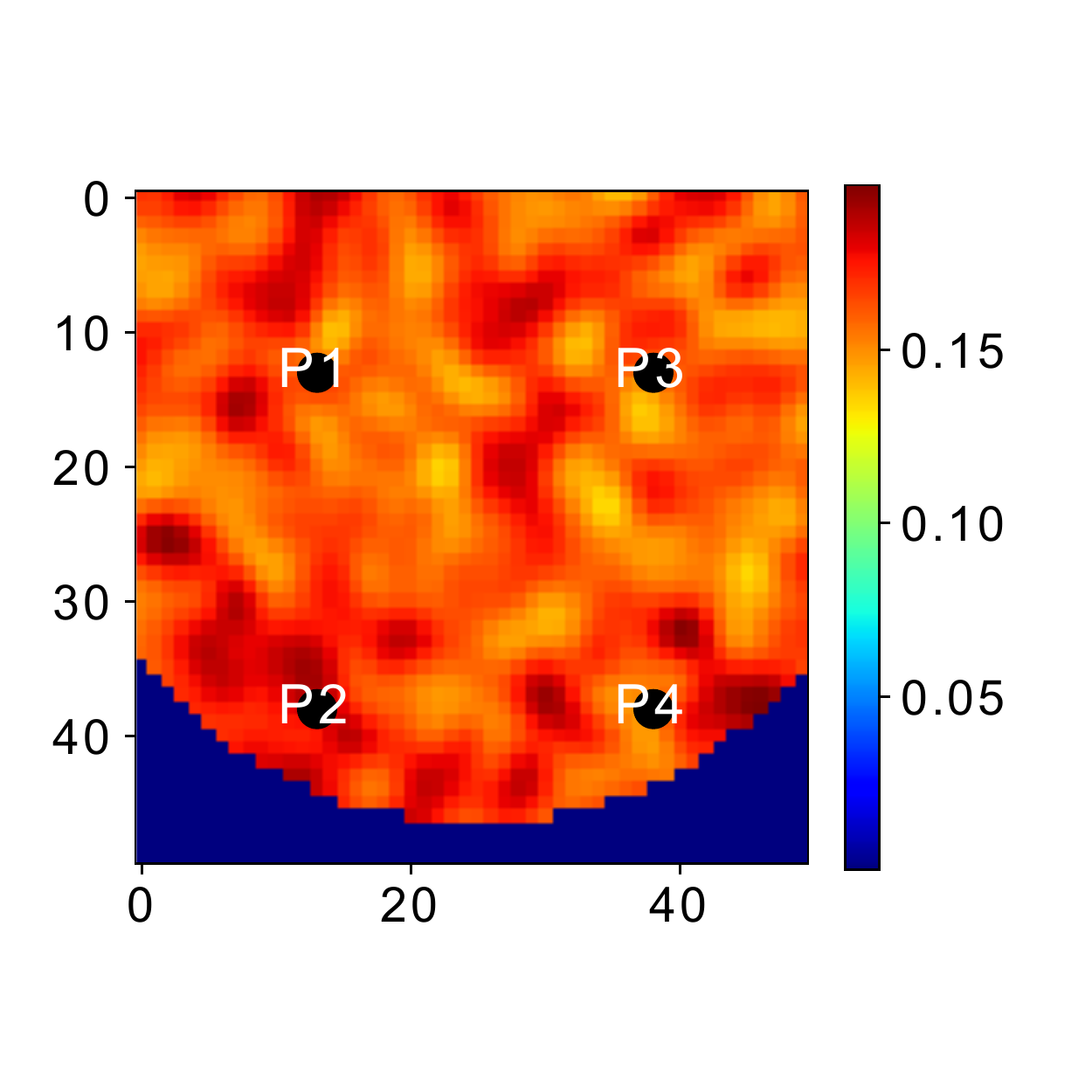}
        \vspace{-3\baselineskip}
        \caption{4-spot}
    \end{subfigure}%
    \hspace{2\baselineskip}
    \begin{subfigure}[b]{0.43\textwidth}
        \includegraphics[width=1\textwidth]{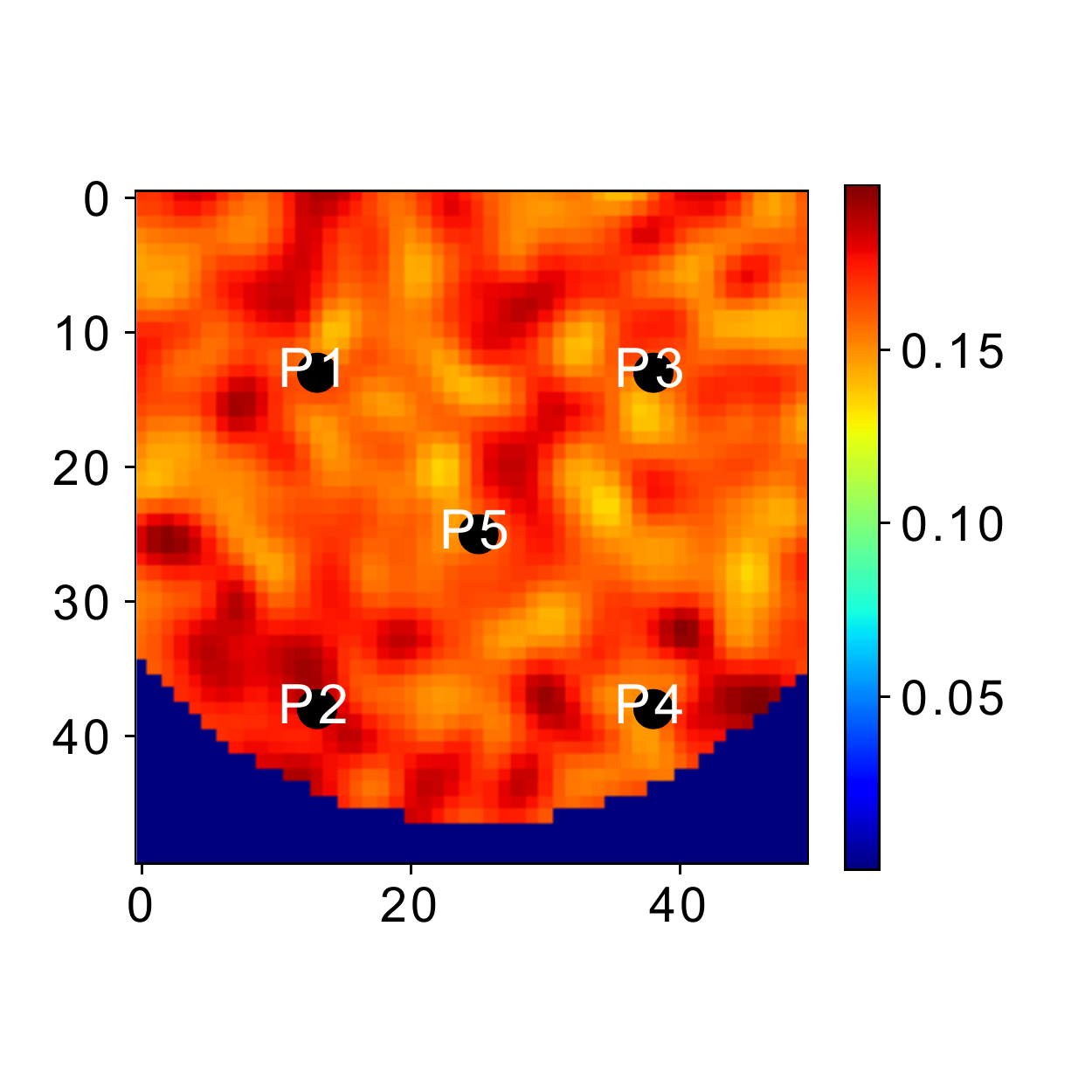}
        \vspace{-3\baselineskip}
        \caption{5-spot}
    \end{subfigure}%
    \vspace{-2\baselineskip}
    \begin{subfigure}[b]{0.43\textwidth}
        \includegraphics[width=1\textwidth]{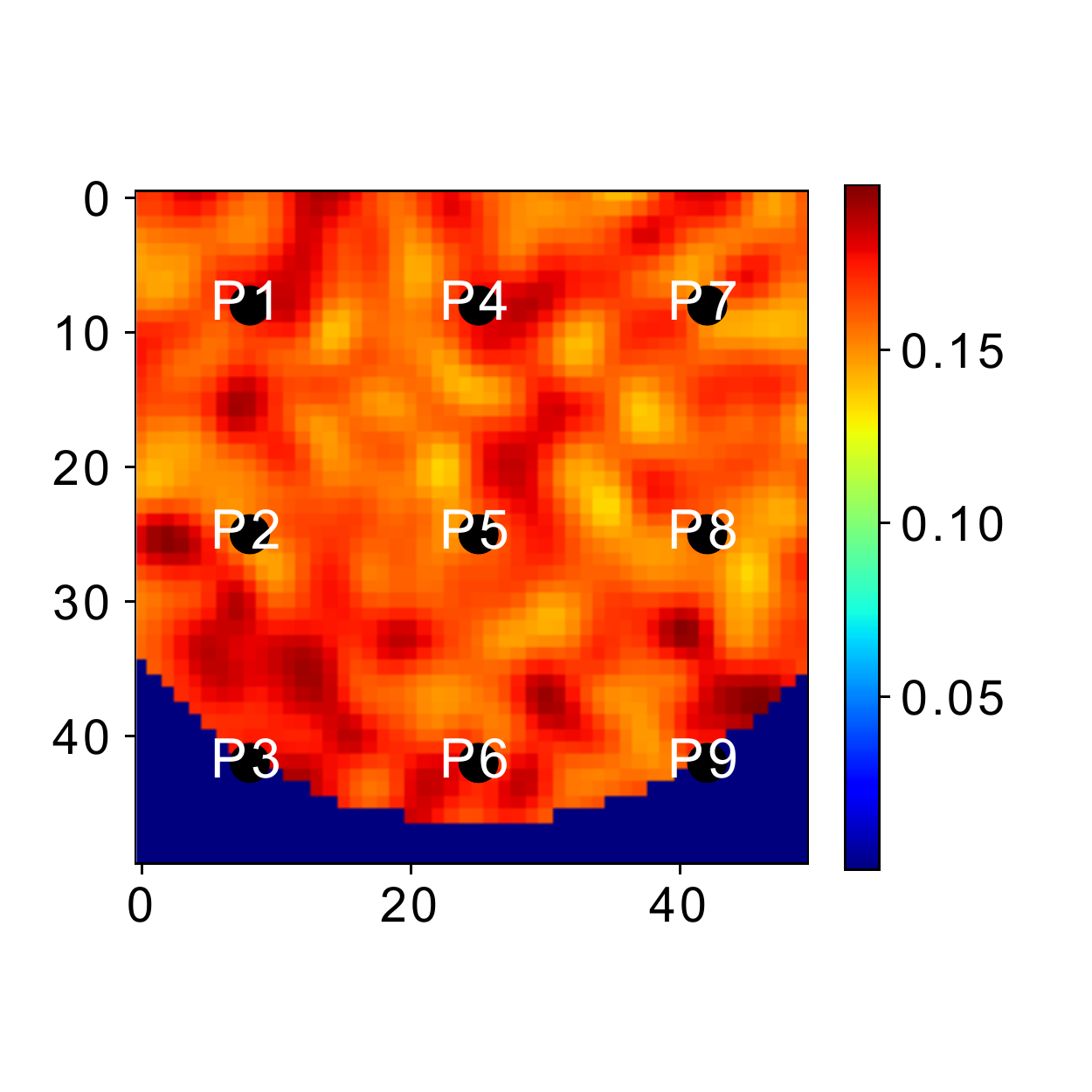}
        \vspace{-3\baselineskip}
        \caption{9-spot}
    \end{subfigure}%
    \hspace{2\baselineskip}
    \begin{subfigure}[b]{0.43\textwidth}
        \includegraphics[width=1\textwidth]{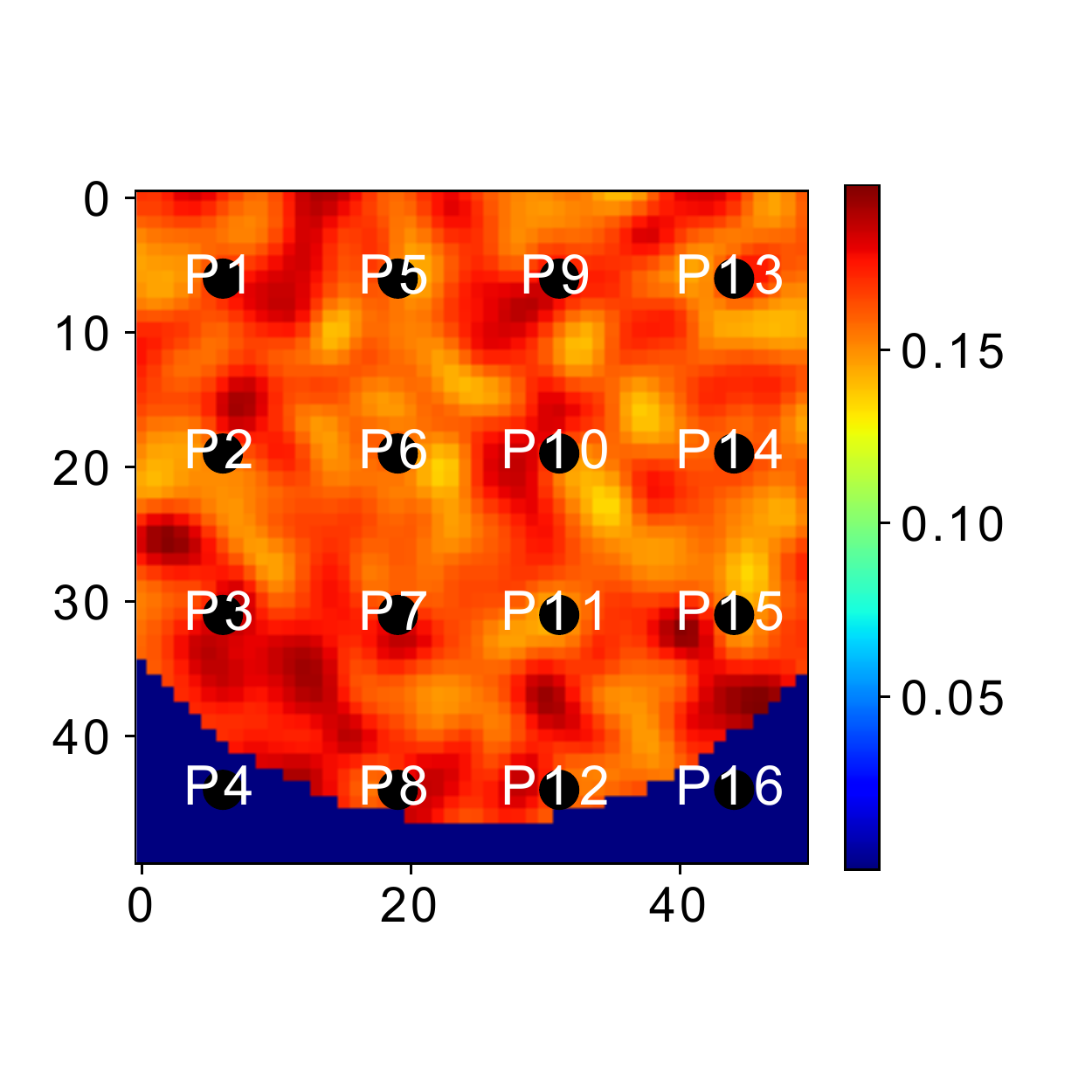}
        \vspace{-3\baselineskip}
        \caption{16-spot}
    \end{subfigure}%

     \caption{Illustration of the 4-, 5-, 9- and 16-spot well pattern with the background as the porosity field of a specific test scenario.}
	\label{fig:well_patterns}
\end{figure}

The performance of the best AI agent found using the dual-action probability distribution is now compared to reference well-pattern agents. The best artificial intelligence agent is taken to be the agent with the maximum average NPV of the 150 field development scenarios previously considered. The reference well patterns considered include the 4-, 5-, 9- and 16-spot patterns which are illustrated in Fig.~\ref{fig:well_patterns}. Note that these patterns are made up of only production wells and the wells are equally spaced. Wells that fall in the inactive region (dark-blue region) of the reservoir are not considered in the field development. For example, wells P3 and P9 in the 9-spot pattern (Fig.~\ref{fig:well_patterns}~(c)) and wells  P4 and P16 in the 16-spot pattern (Fig.~\ref{fig:well_patterns}~(d)) are removed from the development plan.

\begin{figure}[!htb]
    \centering
   
    \begin{subfigure}[b]{0.43\textwidth}
        \includegraphics[width=1\textwidth]{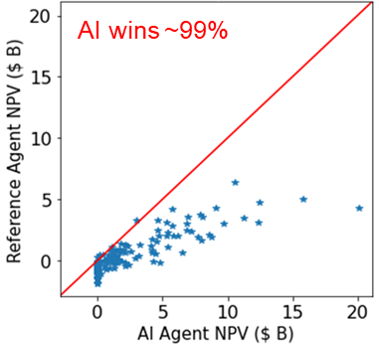}
        \caption{4-spot}
    \end{subfigure}%
    \hspace{2\baselineskip}
    \begin{subfigure}[b]{0.43\textwidth}
        \includegraphics[width=1\textwidth]{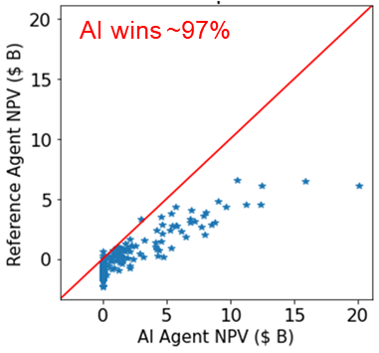}
        \caption{5-spot}
    \end{subfigure}%
    
    \begin{subfigure}[b]{0.43\textwidth}
        \includegraphics[width=1\textwidth]{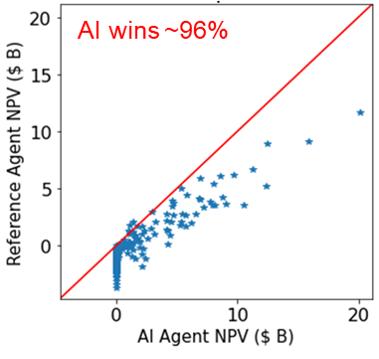}
        \caption{9-spot}
    \end{subfigure}%
    \hspace{2\baselineskip}
    \begin{subfigure}[b]{0.43\textwidth}
        \includegraphics[width=1\textwidth]{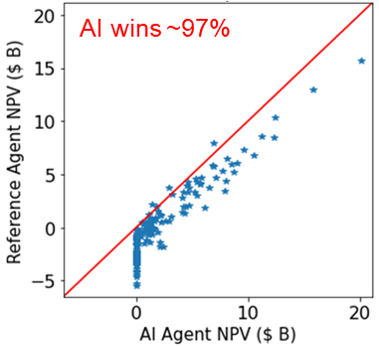}
        \caption{16-spot}
    \end{subfigure}%
    
    \begin{subfigure}[b]{0.45\textwidth}
        \includegraphics[width=1\textwidth]{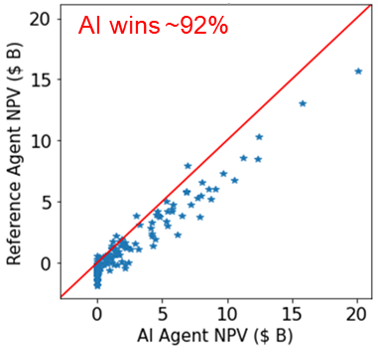}
        \caption{Maximum from all reference agents}
    \end{subfigure}%
    
     \caption{Comparison of the best artificial intelligence (AI) agent trained using the dual-action probability representation with the reference well-pattern agents for the 2D case.}
	\label{fig:comp_ai_pattern}
\end{figure}

Figure~\ref{fig:comp_ai_pattern}~(a),~(b),~(c),~and~(d) shows the performance of the artificial intelligence agent against the 4-, 5-, 9- and 16-spot reference well patterns, respectively. In all four cases, the artificial intelligence agent outperforms the reference well patterns in at least 96\% of the 150 field development scenarios considered.

Figure~\ref{fig:comp_ai_pattern}~(e) shows the comparison of the NPV obtained from the artificial intelligence agent with the maximum NPV obtained from the four well-pattern agents for each of the 150 field development scenarios. In this case, the artificial intelligent agent outperforms the maximum from all well-pattern agents in approximately 92\% of the field development scenarios. The minimum NPV obtained by the artificial intelligence agent in the 150 field development scenarios is zero NPV. For a significant proportion of these cases, the drilling of wells using the well-pattern agents leads to negative NPV, while the AI agent simply recommended not to develop the field. The results demonstrate the ability of the artificial intelligence agent to identify unfavorable field development scenarios. Although not considered in this work, various well spacing, well-pattern geometry, or orientation, such as those discussed in well-pattern optimization~\citep{Onwunalu2011ADevelopment, nasir2021two}, could be used to further refine the performance of the artificial intelligence agent.

\begin{figure}[!htb]
    \centering
    \begin{subfigure}[b]{0.35\textwidth}
        \includegraphics[width=1\textwidth]{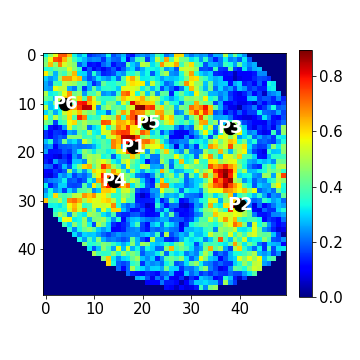}
        \caption{Well index (scenario 1)}
    \end{subfigure}%
    ~
    \begin{subfigure}[b]{0.35\textwidth}
        \includegraphics[width=1\textwidth]{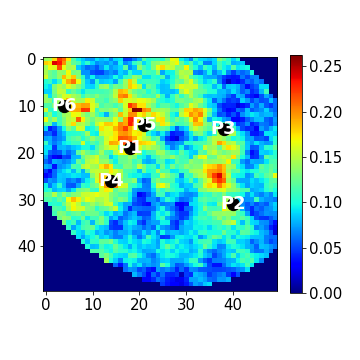}
        \caption{Oil accumulation (scenario 1)}
    \end{subfigure}%
    ~
    \begin{subfigure}[b]{0.35\textwidth}
        \includegraphics[width=1\textwidth]{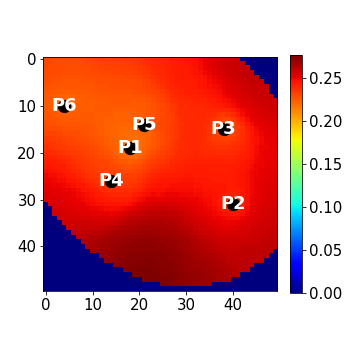}
        \caption{Pressure (scenario 1)}
    \end{subfigure}%
    
    \begin{subfigure}[b]{0.35\textwidth}
        \includegraphics[width=1\textwidth]{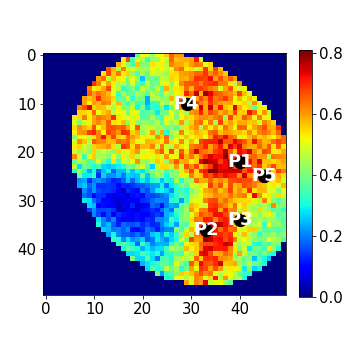}
        \caption{Well index (scenario 2)}
    \end{subfigure}%
    ~
    \begin{subfigure}[b]{0.35\textwidth}
        \includegraphics[width=1\textwidth]{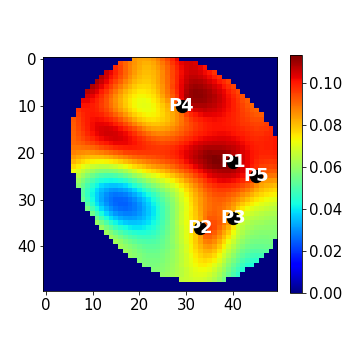}
        \caption{Oil accumulation (scenario 2)}
    \end{subfigure}%
    ~
    \begin{subfigure}[b]{0.35\textwidth}
        \includegraphics[width=1\textwidth]{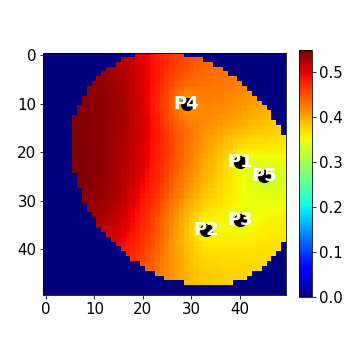}
        \caption{Pressure (scenario 2)}
    \end{subfigure}%
    
    \begin{subfigure}[b]{0.35\textwidth}
        \includegraphics[width=1\textwidth]{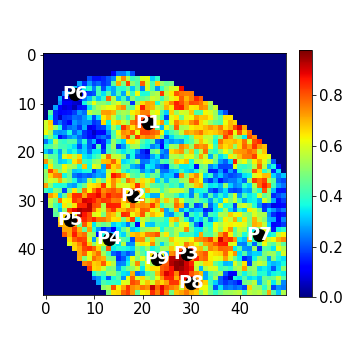}
        \caption{Well index (scenario 3)}
    \end{subfigure}%
    ~
    \begin{subfigure}[b]{0.35\textwidth}
        \includegraphics[width=1\textwidth]{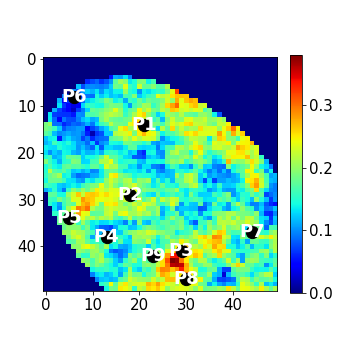}
        \caption{Oil accumulation (scenario 3)}
    \end{subfigure}%
    ~
    \begin{subfigure}[b]{0.35\textwidth}
        \includegraphics[width=1\textwidth]{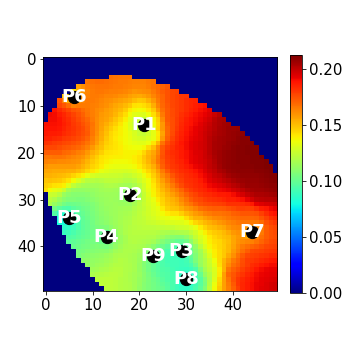}
        \caption{Pressure (scenario 3)}
    \end{subfigure}%

    \caption{Well configuration proposed by the best artificial intelligence agent for three random 2D field development scenarios. The well numbers indicate the drilling stage in which the wells are drilled.}
    \label{fig:well_configs}
\end{figure}

The positioning of wells by the artificial intelligence agent for three random cases out of the 150 field development scenarios are shown in Fig.~\ref{fig:well_configs}. These cases entails field development plans with six, five, and nine production wells, in scenario 1, 2, and 3, respectively. All wells in the three scenarios are drilled early. For example, in scenario 1, the six wells are drilled in the first six drilling stages with well P1 drilled in the first drilling stage and P6 in the sixth drilling stage. The wells are strategically placed in regions of the reservoir with high productivity (Fig.~\ref{fig:well_configs}~(a)~(d)~(g)) and oil accumulation (Fig.~\ref{fig:well_configs}~(b)~(e)~(h)). The pressure distributions for the scenarios are shown in Fig.~\ref{fig:well_configs}~(c)~(f)~(i). The wells are also properly spaced resulting in a good coverage of the producing region.

\subsection{Case 2: Three-dimensional system}
\label{sec:3D}

Results from the training of the agent for the 3D subsurface system is now presented. For the 3D case, the wells are assumed to be always vertical and fully penetrate all layers of the reservoir in this work. The procedures discussed in this work can, however, be readily extended to horizontal wells with varying completion strategies. Compared to that of the 2D case, a larger  neural network (in terms of number of learnable parameters) is used to represent the policy for the 3D case. This is because the size of the input channels to the agent is larger than that of the 2D case. The vertical heterogeneity in the reservoir models is captured by including the transmissibility in the Z-direction in the state representation.

\begin{figure}[!htb]
    \centering
   
    \begin{subfigure}[b]{0.45\textwidth}
        \includegraphics[width=1.1\textwidth]{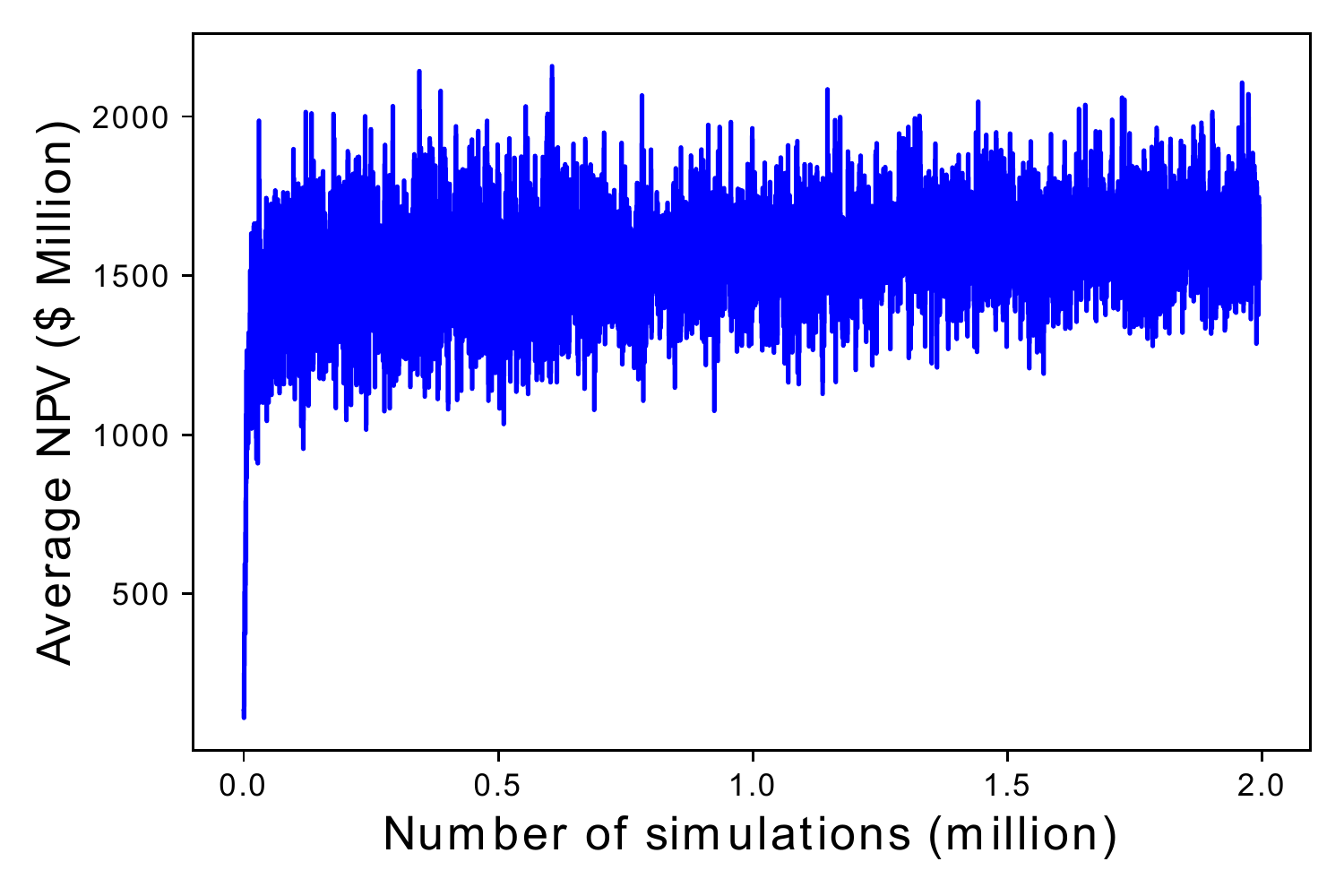}
        \caption{Average NPV}
    \end{subfigure}%
    \hspace{1\baselineskip}
    \begin{subfigure}[b]{0.45\textwidth}
        \includegraphics[width=1.1\textwidth]{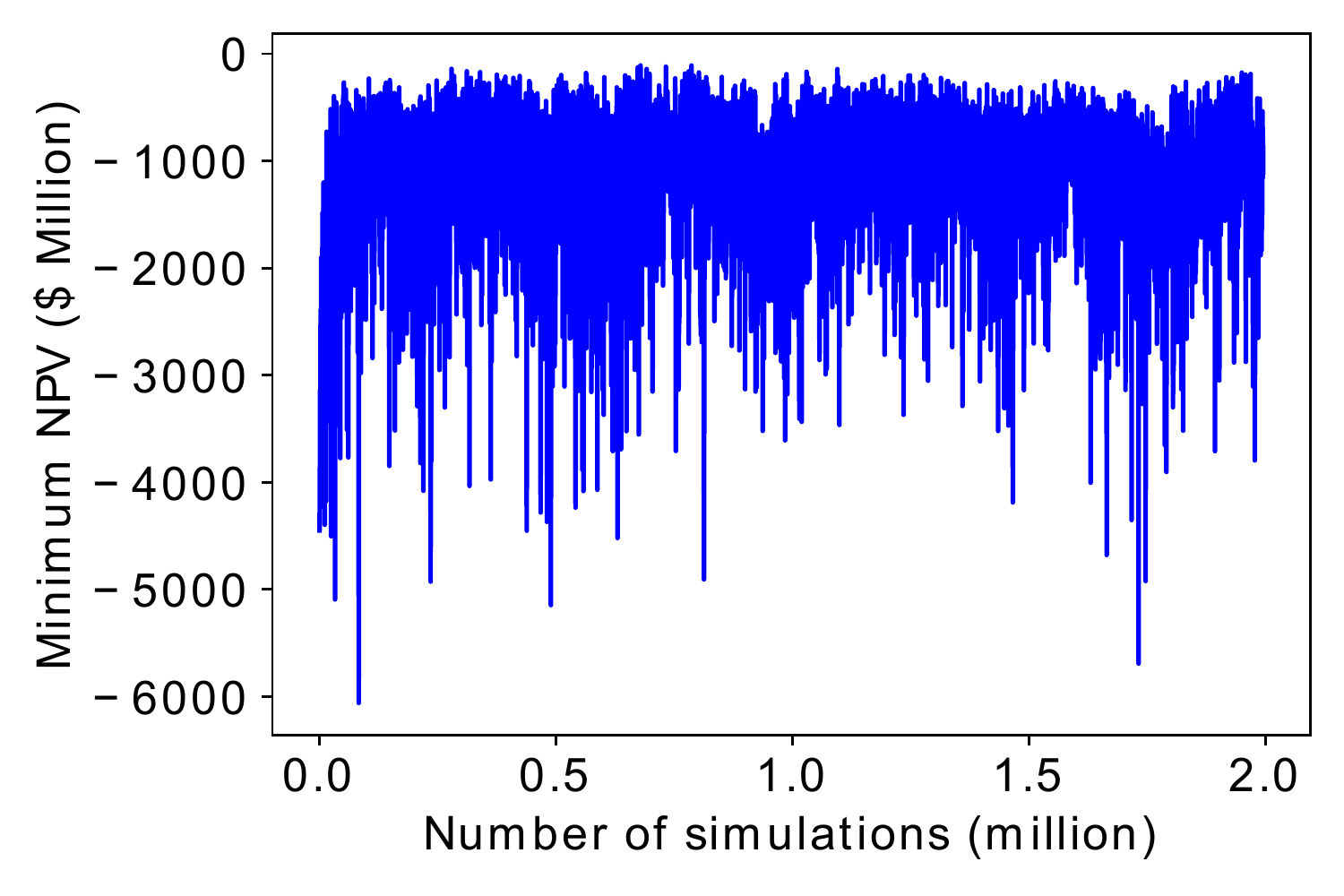}
        \caption{Minimum NPV}
    \end{subfigure}%
    
    \begin{subfigure}[b]{0.45\textwidth}
        \includegraphics[width=1.1\textwidth]{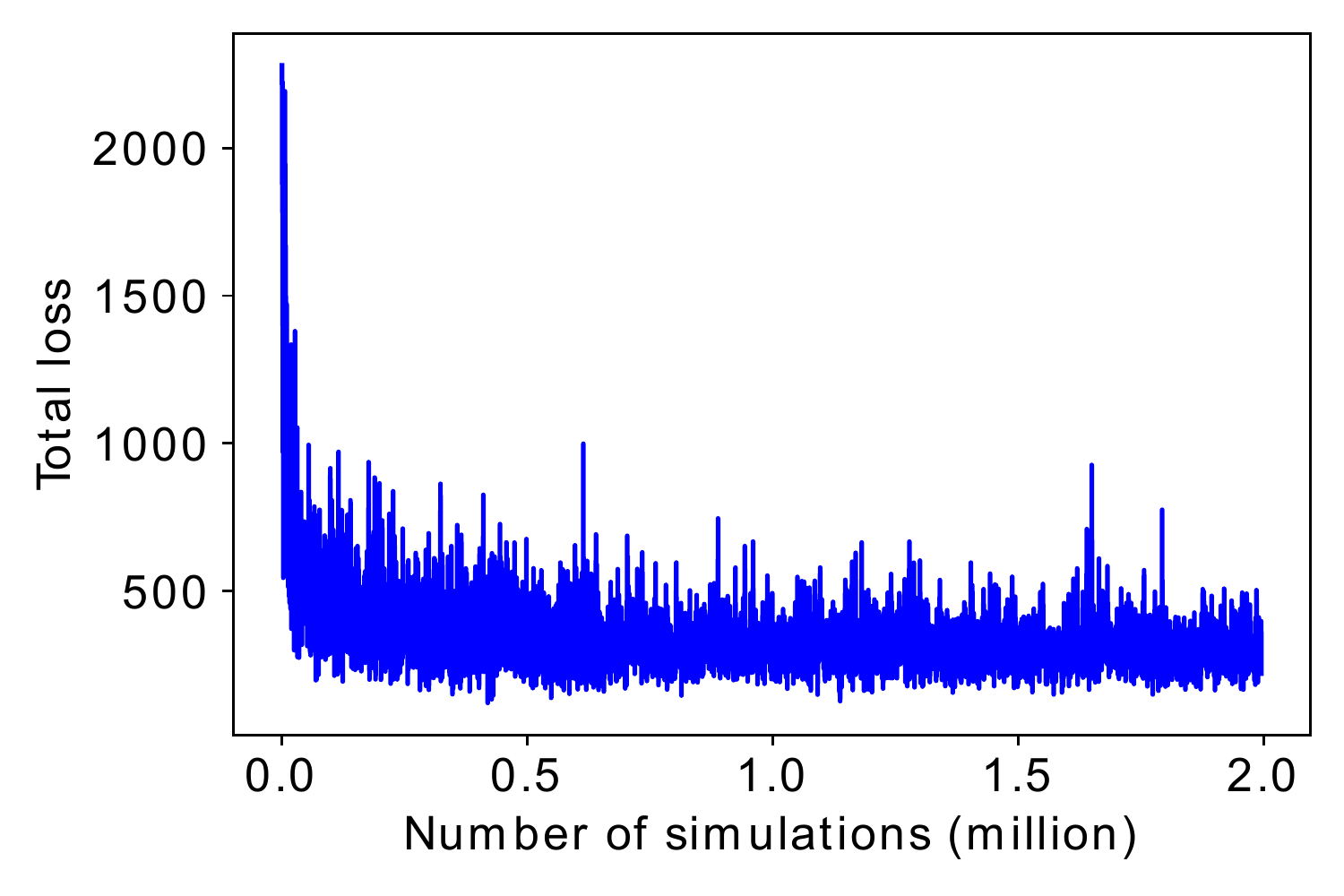}
        \caption{Total loss}
    \end{subfigure}%
    \hspace{1\baselineskip}
    \begin{subfigure}[b]{0.45\textwidth}
        \includegraphics[width=1.1\textwidth]{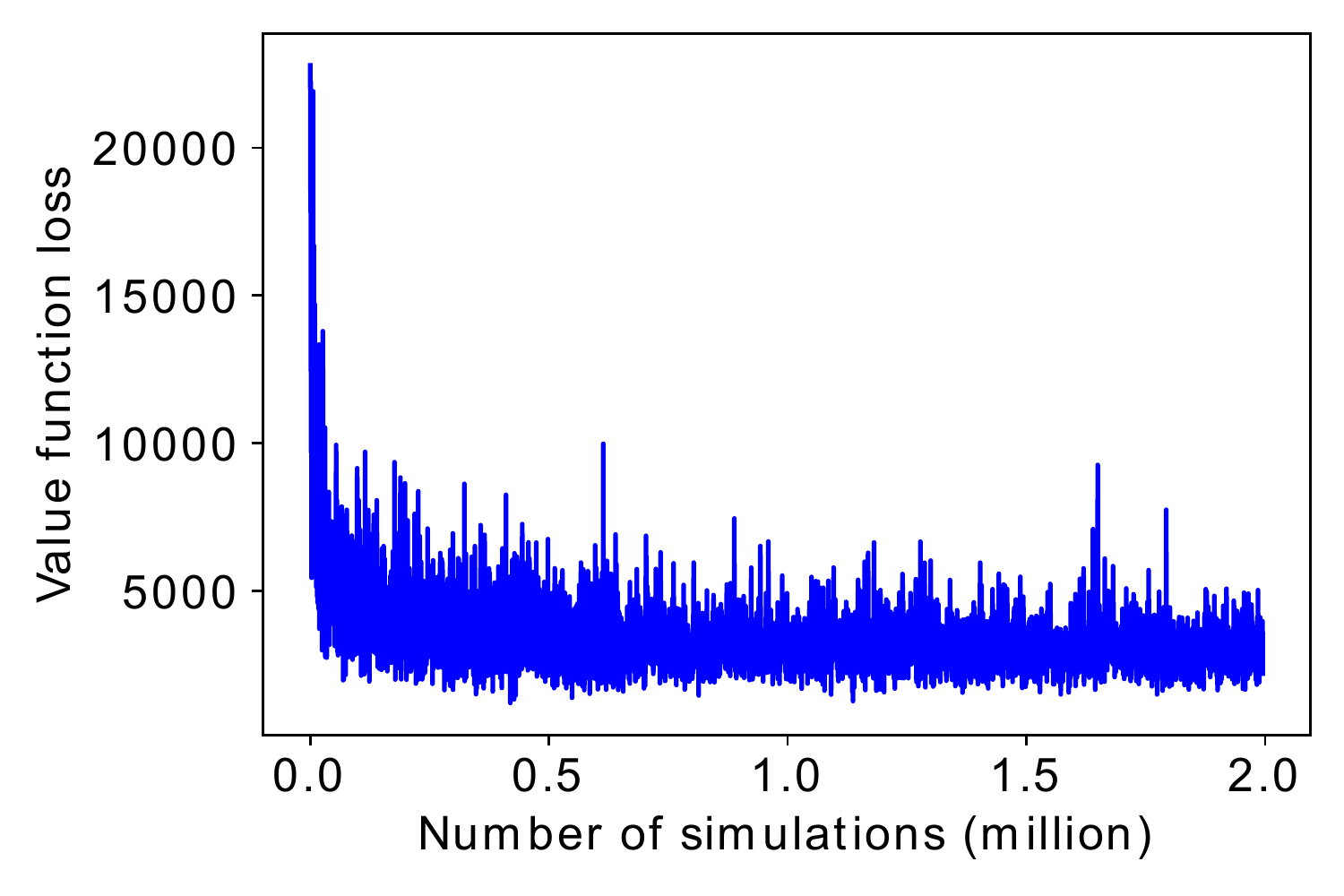}
        \caption{Value function loss}
    \end{subfigure}%
    
    \begin{subfigure}[b]{0.45\textwidth}
        \includegraphics[width=1.1\textwidth]{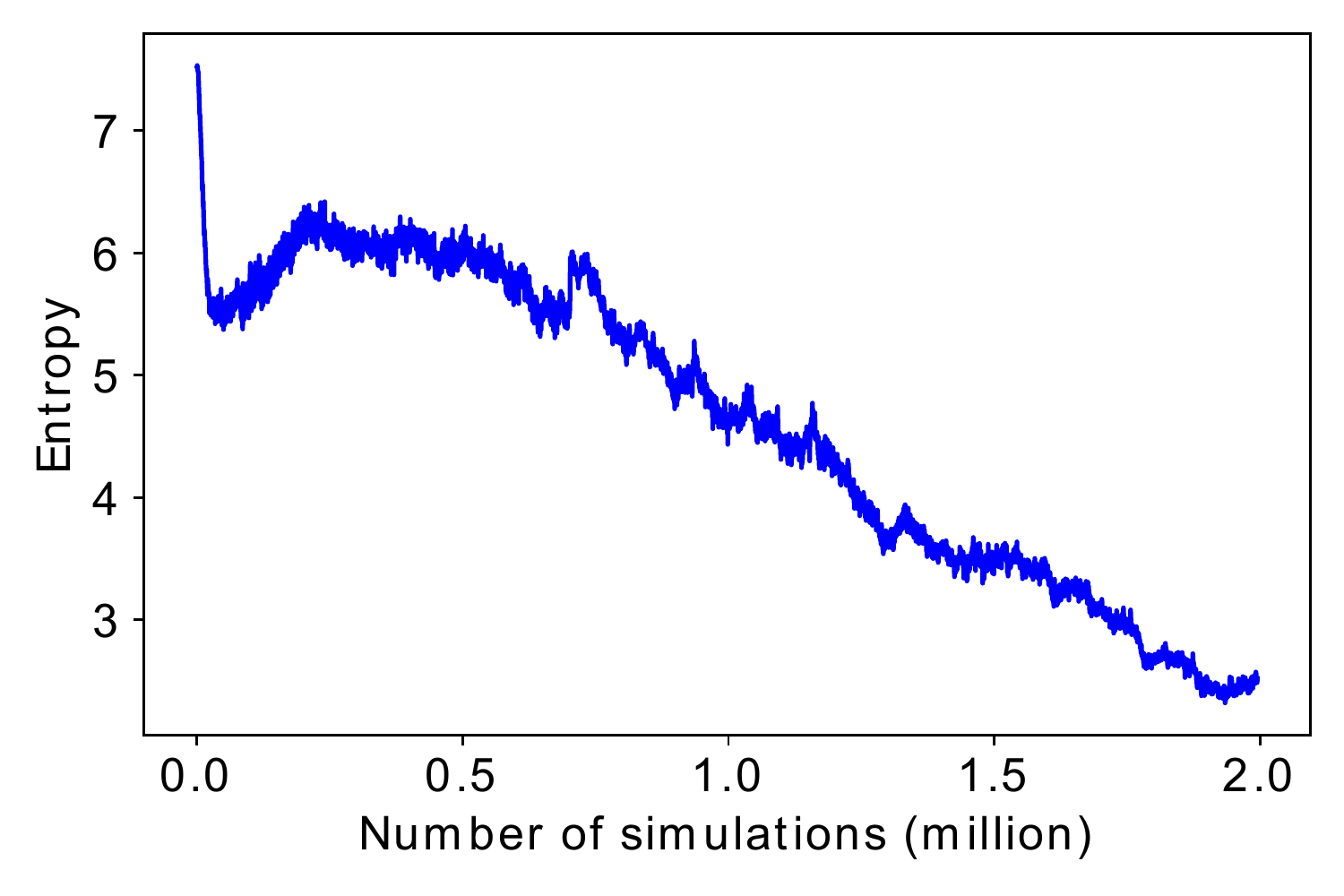}
        \caption{Entropy}
    \end{subfigure}%
    \hspace{1\baselineskip}
    \begin{subfigure}[b]{0.45\textwidth}
        \includegraphics[width=1.1\textwidth]{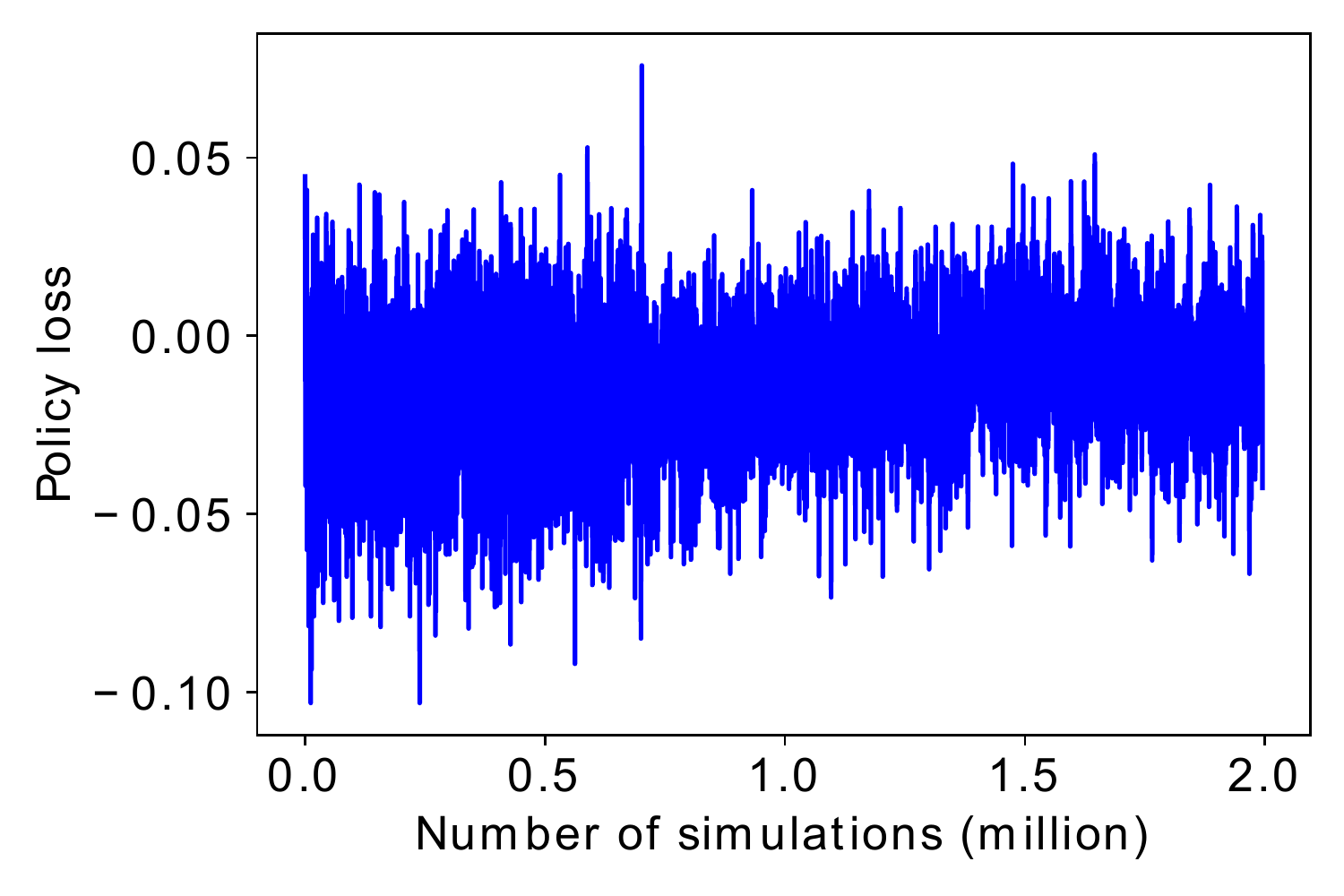}
        \caption{Policy loss}
    \end{subfigure}%
    
     \caption{Evolution of training performance metrics for the dual-action probability distributions for the 3D case.}
	\label{fig:kpi_cur_param_3d}
\end{figure}

Figure~\ref{fig:kpi_cur_param_3d} shows the evolution of the performance indicators for the training process using the dual-action probability distribution for the 3D case. The training entails approximately 2 million simulations. From Fig.~\ref{fig:kpi_cur_param_3d}~(a) it can be seen that the average NPV generally increases as the training progresses. This demonstrates the artificial intelligence agent can learn the field development logic for the even more complicated three-dimensional system. Consistent with the behaviour of the agent for the two-dimensional case when the dual-action probability distribution is used, we see a rapid increase in average NPV during the initial phase of the training. As observed in the 2D case, the total loss (Fig.~\ref{fig:kpi_cur_param_3d}~(c)), value function loss (Fig.~\ref{fig:kpi_cur_param_3d}~(d)), and entropy loss (Fig.~\ref{fig:kpi_cur_param_3d}~(e)) decrease as the training progresses.

\begin{figure}[!htb]
    \centering
    % INIT
        
        \floatbox[{\capbeside\thisfloatsetup{capbesideposition={left,top},capbesidewidth=7.2cm}}]{figure}[\FBwidth]
{\caption{Comparison of the best artificial intelligence (AI) agent trained using the dual-action probability representation with the maximum NPV from the reference well-pattern agents for the 3D case. }\label{fig:comp_3d_ai_pat}{\hspace{0.2\baselineskip}}}
    {\includegraphics[width=0.55\textwidth]{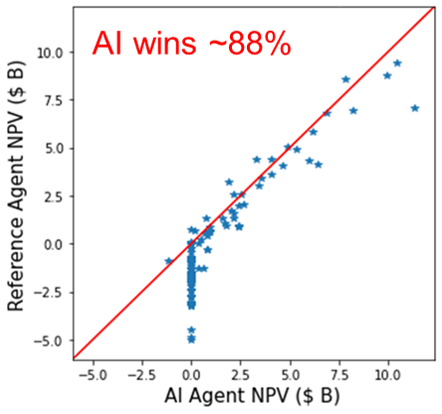}}
\end{figure}

The performance of the best artificial intelligence agent is also compared with the four reference well-pattern agents for a set of random field development scenarios. The performance of the artificial intelligence agent against the maximum NPV obtained from the four well-pattern agents (same as those considered in the 2D case) is shown in Fig.~\ref{fig:comp_3d_ai_pat}. The artificial intelligence agent outperforms the well-pattern agents in approximately 88\% of the cases considered.

\begin{figure}[!htb]
    \centering
    \begin{subfigure}[b]{0.34\textwidth}
        \includegraphics[width=1\textwidth]{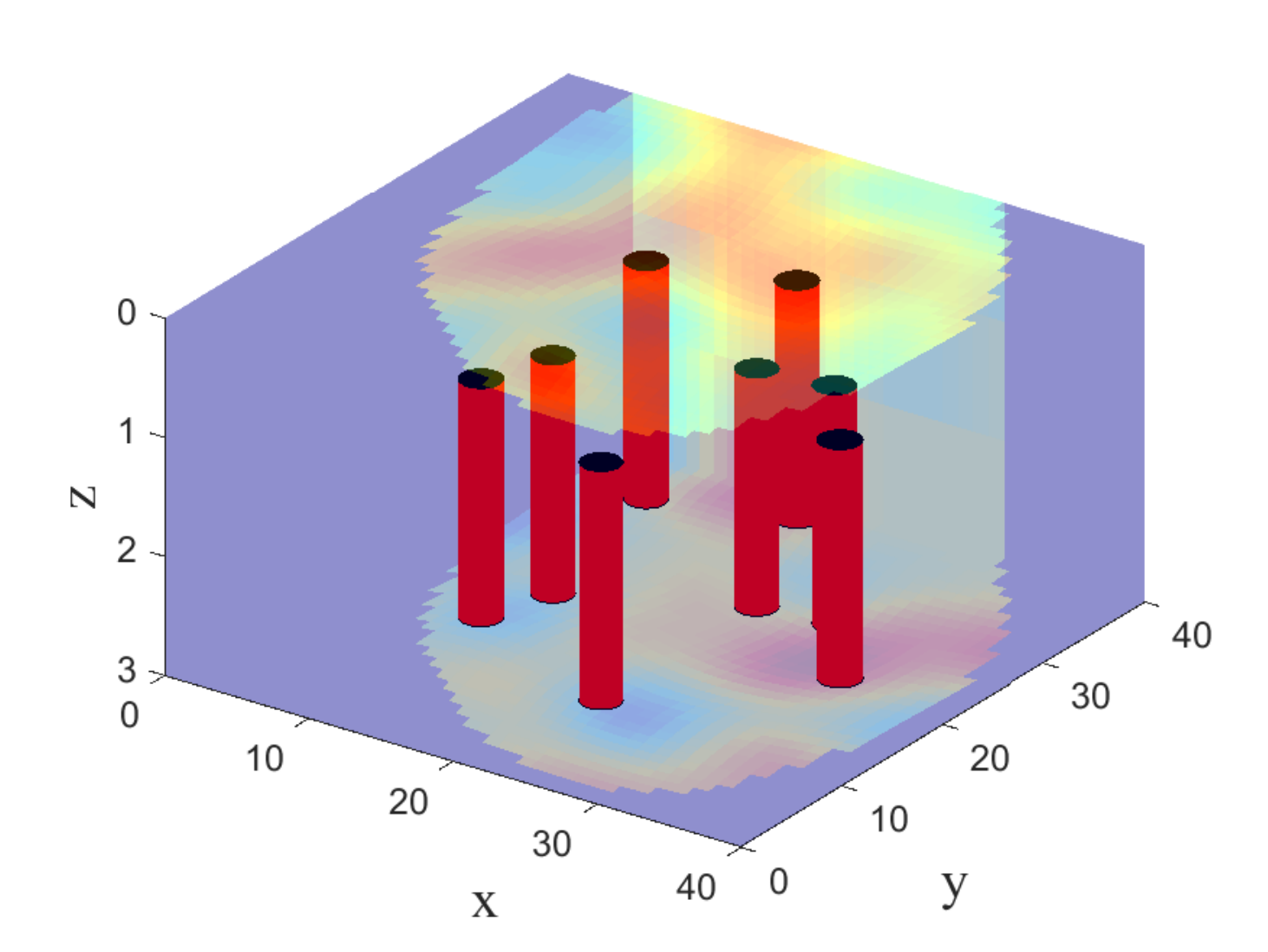}
        \caption{scenario 1}
    \end{subfigure}%
    ~
    \begin{subfigure}[b]{0.34\textwidth}
        \includegraphics[width=1\textwidth]{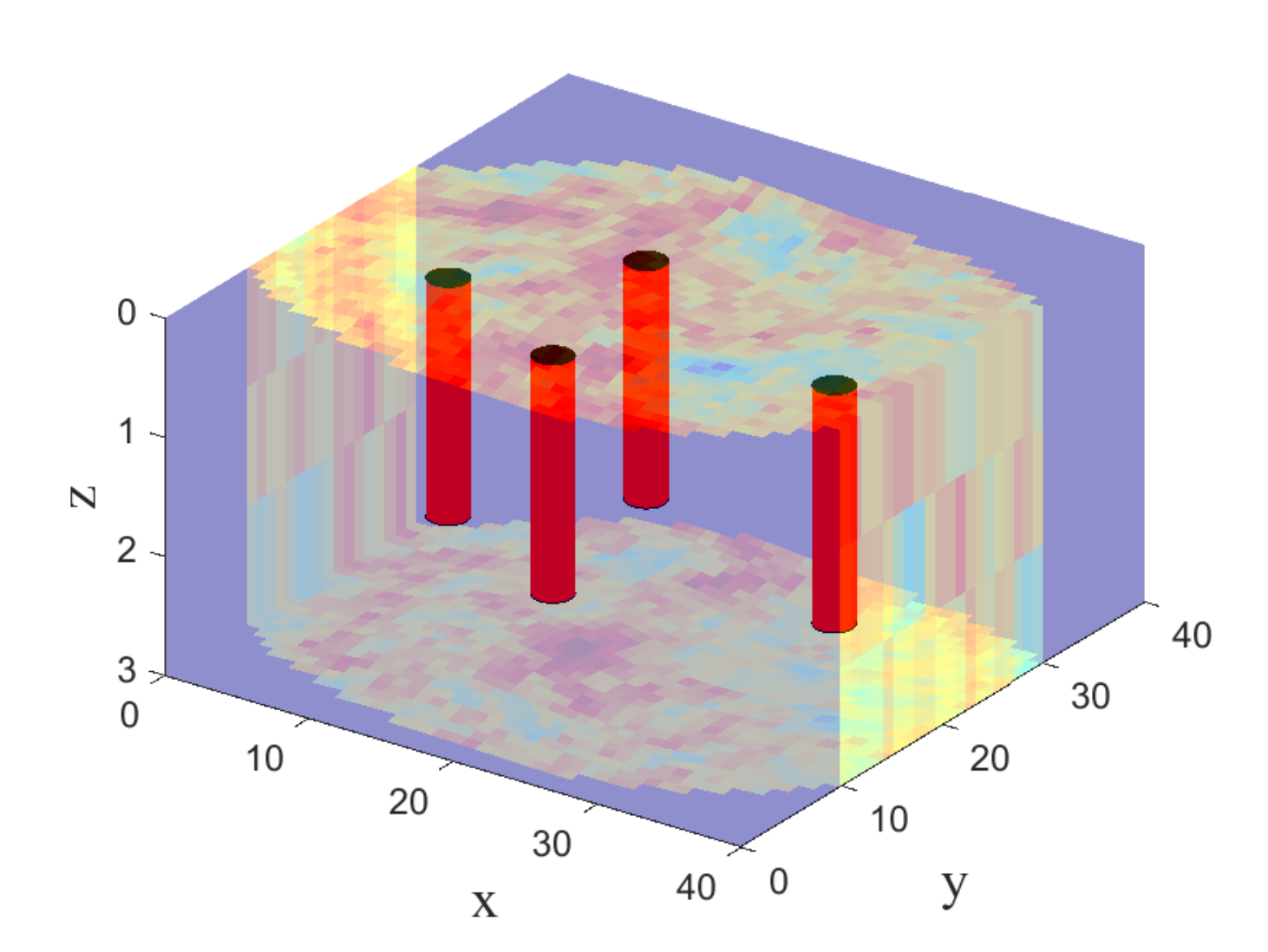}
        \caption{scenario 2}
    \end{subfigure}%
    ~
    \begin{subfigure}[b]{0.35\textwidth}
        \includegraphics[width=1\textwidth]{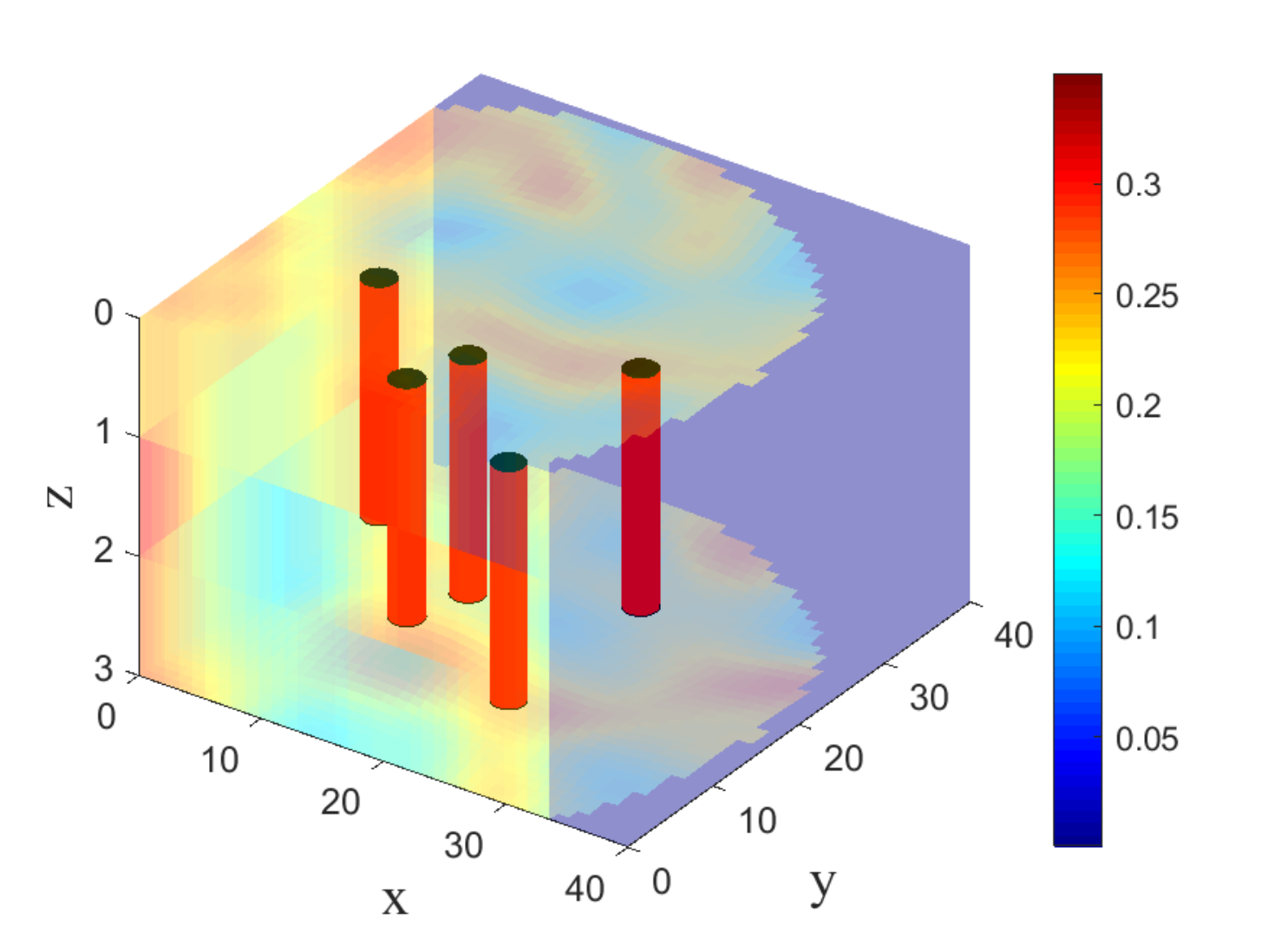}
        \caption{scenario 3}
    \end{subfigure}%

    \caption{Well configuration proposed by the best artificial intelligence agent for three random 3D field development scenarios.}
    \label{fig:well_configs_3D}
\end{figure}

Figure~\ref{fig:well_configs_3D} shows the well configurations proposed by the best AI agent for three random field development scenarios. The field development plans involves eight, four and five production wells. 

\section{Concluding Remarks}
\label{sec:concl}

In this work, we developed an AI agent for constrained field development optimization for two- and three-dimensional subsurface two-phase flow models. The training of the agent utilizes the concept of deep reinforcement learning where a feedback paradigm is used to continuously improve the performance of the agent through experience. The experience includes the action or decision taken by the agent, how this action affects the state of the environment (on which the action is taken), and the corresponding reward which indicates the quality of the action. The training efficiency is enhanced using a dual-action probability distribution parameterization and a convolutional neural network architecture with shared layers for the policy and value functions of the agent.  After appropriate training, the agent instantaneously provides optimized field development plan, which includes the number of wells to drill, their location, and drilling sequence for different field development scenarios within a predefined range of applicability.

Example cases involving 2D and 3D subsurface systems are used to assess the performance of the training procedure and the resulting artificial intelligence agent. The use of the dual-action probability distribution shows clear advantage over the single-action probability distribution for the training of the artificial intelligence agent. The trained artificial intelligence agents for the 2D and 3D case are shown to outperform four reference well-pattern agents. For the 2D case, the artificial intelligence agent found a better field development plan than the reference agents in approximately 92\% of 150 random field development scenarios. The trained agent for the 3D case outperformed the reference well-pattern agents in approximately 88\% of a set of random field development scenarios. The results demonstrated the ability of the trained agents to avoid developing unfavorable field development scenarios and strategically place wells in regions of high productivity.

In future work, the performance of other deep reinforcement learning algorithms, such as soft actor critic (SAC)~\citep{haarnoja2018soft}, importance weighted actor-learner architectures (IMPALA)~\citep{espeholt2018impala} (and PPO variant of IMPALA), that have demonstrated great sample and computational efficiency in other domains should be evaluated for the field development optimization problem. The proposed procedure should also be extended to the joint optimization of well locations and operational settings. This would require an additional variable in the action to represent the operation settings of each well. Finally, the deep reinforcement learning framework should be extended to other field development cases such as waterflooding and optimization under uncertainty where multiple realizations of the geological model are used to represent geological uncertainty.

\begin{acknowledgements}

We thank Chevron Technical Center for permission to publish this work. We would also like to thank the Reinforcement Learning Team in Microsoft who provided support during this project. Finally, we thank Denis Voskov and the Delft Advanced Research Terra Simulator (DARTS) team at TU Delft for providing the reservoir simulator used in this work.
\end{acknowledgements}

\bibliographystyle{spbasic}      % basic style, author-year citations
\bibliography{DRL}

\end{document}